\documentclass[graybox]{svmult}

\usepackage{imakeidx} 

\usepackage{type1cm}        % activate if the above 3 fonts are
\usepackage{makeidx}         % allows index generation
\usepackage{graphicx}        % standard LaTeX graphics tool
\usepackage{multicol}        % used for the two-column index
\usepackage[bottom]{footmisc}% places footnotes at page bottom

\usepackage{newtxtext}       % 
\usepackage{newtxmath}       % selects Times Roman as basic font

\usepackage{syntax}
\usepackage{pgf}
\usepackage{tikz}
\usetikzlibrary{positioning,arrows.meta}
\usepackage{subcaption}
\usepackage{rotating}

\definecolor{exploration}{gray}{0.8}
\newenvironment{exploration}[1]{\ignorespaces\def\stmtopen##1{##1}%
\formtmp{exploration}{\ \circledmark{\textcolor{black}{$~\blacktriangleright$}}{\kern6pt}#1}}{\par\noindent\textcolor{exploration}{\rule{\columnwidth}{1pt}}\vskip2pt\par\addvspace{\baselineskip}}%

\makeindex             % used for the subject index
\begin{document}

\title*{A Guide to Large Language Models in Modeling and Simulation: From Core Techniques to Critical Challenges}
\titlerunning{Large Language Models for Modeling \& Simulation}

\author{Philippe J. Giabbanelli}

\institute{Philippe J. Giabbanelli \at VMASC, Old Dominion University, USA, \email{pgiabban@odu.edu}}


\maketitle

\abstract{Large language models (LLMs) have rapidly become familiar tools to researchers and practitioners. Concepts such as prompting, temperature, or few-shot examples are now widely recognized, and LLMs are increasingly used in Modeling \& Simulation (M\&S) workflows. However, practices that appear straightforward may introduce subtle issues, unnecessary complexity, or may even lead to inferior results. Adding more data can backfire (e.g., deteriorating performance through model collapse or inadvertently wiping out existing guardrails), spending time on fine-tuning a model can be unnecessary without a prior assessment of what it already knows, setting the temperature to 0 is not sufficient to make LLMs deterministic, providing a large volume of M\&S data as input can be excessive (LLMs cannot attend to everything) but naive simplifications can lose information. We aim to provide comprehensive and practical guidance on how to use LLMs, with an emphasis on M\&S applications. We discuss common sources of confusion, including non-determinism, knowledge augmentation (including RAG and LoRA), decomposition of M\&S data, and hyper-parameter settings. We emphasize principled design choices, diagnostic strategies, and empirical evaluation, with the goal of helping modelers make informed decisions about when, how, and whether to rely on LLMs.}


%% The outline of the book chapter
\section{Introduction}
\begin{figure}[t]
\centering
\includegraphics[width=\textwidth]{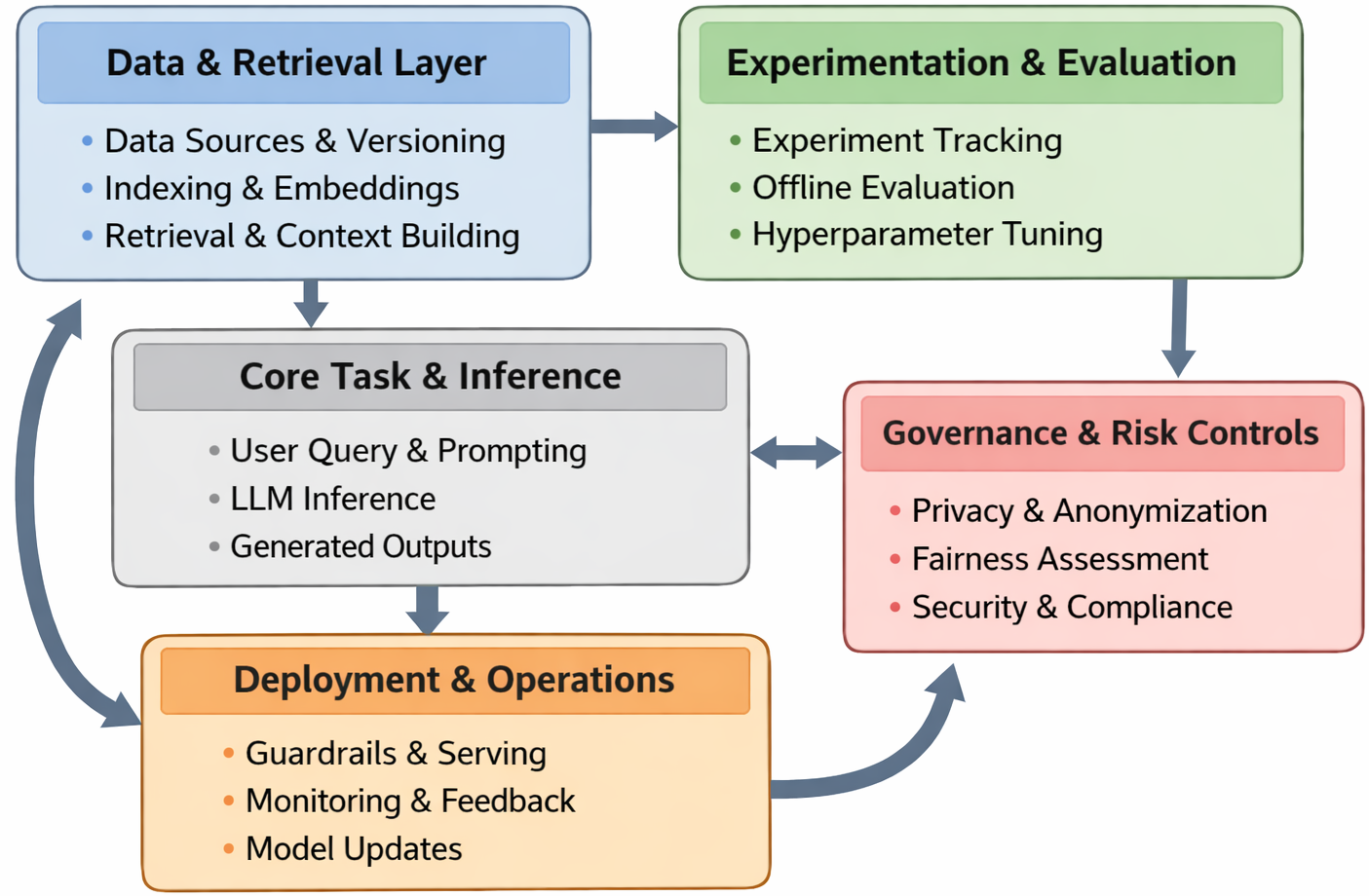}
\caption{
Rather than a single prompt–response interaction, LLM-based systems are organized as \textit{pipelines}: we need to know what version of the data is used (e.g., via {\ttfamily data version control}), automate the experiments (e.g., through {\ttfamily MLflow}), ensure that personally identifiable information is detected and removed (e.g., using {\ttfamily Presidio}), evaluate with respect to fairness, etc. \textit{Orchestration}\index{Orchestration} (e.g., based on {\ttfamily LangChain}) is necessary to coordinate these many aspects.
}
\label{fig:LLMpipelines}
\end{figure}

Generative AI (GenAI) is often associated with Large Language Models (LLMs), although the field also covers image or video generation. In the early 2020s, there was a sense of marvel (or disappointment) depending on whether an LLM could produce the right textual output solely based on prompts and a series of hand-curated examples, using either simple notebooks in Python or a web-portal. Prompting could already be challenging as studies showed that users relied on trial-and-error~\cite{zamfirescu2023johnny,ma2025should}, with researchers noting that ``finding the right prompt or model has become an industry unto itself''~\cite{arawjo2024chainforge}. The LLM ecosystem has only gotten more complex, within a short amount of time that makes it challenging to keep up with technological change. We now design systems with LLMs as part of \textit{pipelines}\index{LLM pipelines|(} that operate in multiple stages involving evaluation, deployment and monitoring with respect to governance and risk controls (Figure~\ref{fig:LLMpipelines}) -- each stage needing its own libraries. As illustrated by other books~\cite{Kamath2024LLMDeepDive}, there are now over 100 `techniques' to optimize the use of LLMs, many `strategies' to handle biases or privacy, and hundreds of benchmarks. This proliferation makes it difficult for users to know the right procedures when using LLMs in their application context. This affects modeling and simulation experts as well as model commissioners and simulation end-users, who may harbor beliefs on LLMs that are disconnected from technical feasibility, reducing them to a magic wand that can `quickly' be used~\cite{xu-etal-2025-language,steyvers2025large}. Our previous observation on infusing Modeling \& Simulation (M\&S) with machine learning can thus be updated as follows:
\begin{quotation}
``when a new approach with high potential is identified, there can be a temptation for end-users to aim at becoming experts in the approach. This is certainly attractive from a logistical standpoint, as end-users can then take care of their own needs, at their pace. However, acquiring another expertise can be challenging for simulationists. [...] Simulationists should be exposed to what [an LLM] can do for them, and what it needs (e.g., data requirements, computational costs). This would position simulationists as informed end-users who can identify reasonable questions in a given context.''~\cite{giabbanelli2019solving}
\end{quotation}

We aim to inform users of LLMs rather than AI/ML experts about the tools and approaches. We draw extensively on our years of experience in using LLMs for M\&S along with the mistakes that we have (unintentionally) made, so that our journey can be of benefit to others. In line with other tutorials and case studies on LLMs for M\&S~\cite{akhavan2024generative,frydenlund2024modeler,vanhee2025large}, Section~\ref{sec:core} begins by covering fundamental components: prompting, hyper-parameters like temperature, and extending the knowledge of LLMs through approaches such as retrieval-augmented generation. Rather than abstract best practices, we examine each of these components as engineering decisions where choices have real consequences for the performance, replicability, or transparency of downstream modeling and simulation tasks. We thus pay particular attention to failures and lack of optimization through reliance on defaults, so that simulation practitioners can reason critically about when and how LLM-based techniques should be employed.

The next sections cover common mistakes and flawed beliefs. Section~\ref{sec:ignoreNonDeterminism} begins with the important matter of non-determinism: we clarify that there are different sources for non-determinism (so it does not just go away by setting the temperature to 0) and we explain how the impact can first be evaluated and, when needed, mitigated. Section~\ref{sec:misuse} covers the temptation of using LLMs because they are a convenient one-stop shop for everything, at the risk of reinventing the wheel or coming up with solutions that appear satisfactory, but are suboptimal. Section~\ref{sec:LLMdiscuss} discusses emerging notions, such as forgetting in LLMs so that we can identify and remove errors instead of adding patches (i.e., shifting from designing LLM pipelines by addition to designing by subtraction), or the potential to use images alongside text as input. Finally, we provide a series of exercises and practical exploration questions that cover decomposing a problem so that an LLM can handle it (instead of simply feeding massive inputs or haphazardly chopping them and losing information), assessing what an LLM already knows, and augmenting its knowledge when needed.

There are many other relevant topics for LLMs, either in general or applied to M\&S in particular, such as comparing with state-of-the-art-solutions, or systematically removing components of the system to understand each part's contribution to overall performance (i.e., performing an \textit{ablation study}\index{Ablation study}). These are common and well motivated requests when reviewing a research paper on LLMs. 

\section{Core elements of LLMs: prompts, parameters, augmentation}
\label{sec:core}
\subsection{Prompt engineering}
The prompt is the textual input provided to an LLM. It is the central element of the LLM-system interface through which users specify tasks, constraints, and evaluation criteria. Since there are several surveys, frameworks~\cite{torkestani2025inclusive}, catalogs of templates~\cite{schulhoff2024prompt,schmidt2024towards}, and taxonomies on prompt engineering (c.f. Figure 2 in~\cite{sahoo2024systematic}), we briefly summarize best principles and then focus on prompt construction as it pertains to modeling and simulation. The highly cited study ``Why Johnny Can't Prompt'' showed that non-experts struggled with generating prompts and evaluating their effectiveness~\cite{zamfirescu2023johnny}. These issues partly stem from defining tasks implicitly, being overconfident about the results, evaluating results with just one metric, debugging by adding more data. In contrast, best practices followed by experts include \textit{defining tasks explicitly}, cautiously \textit{verifying the results}, \textit{systematically debugging}, and \textit{measuring performances through multiple indicators}. Empirical studies have also shown that increasing prompt length or combining multiple prompt engineering techniques does not always improve performance: it may even degrade output quality~\cite{memmert2024more}. For example, the notion of \textit{over-prompting}\index{Prompt engineering!Over-prompting} shows that performance can decrease as we add too many examples~\cite{tang2025few} -- which runs counter to common practices observed in early applications of LLMs. Early studies suggested that the degradation of performances from longer prompts was a problem of retrieval as LLMs may be unable to find information in longer prompts, but newer studies show that \textit{performance degrades with increasing input length even when retrieval is perfect}~\cite{du-etal-2025-context}. Effective prompt construction therefore requires \textit{selectivity} rather than including all information. Overly complex prompts may introduce ambiguity, for example by blurring the distinction between instructions, examples, and counter-examples.

In our experience using LLMs for modeling and simulation, six aspects have been particularly helpful. First, \textbf{the task should be decomposed}\index{Prompt engineering!Task decomposition}. For example, extracting a conceptual model for a corpus is not a specific granular task but a high-level goal. Decomposing it into a series of prompts\footnote{\textit{Agentic AI}\index{Agentic AI} or \textit{LLM multi-agent system} are sometimes used as an umbrella term for workflows with multiple steps~\cite{yuksel2025multi}, such as observe/reason/decide~\cite{xia2024llm}. Each prompt is then called an `agent', so our sample prompts 1 to 3 could be rebranded as agents 1 to 3. We do not use this terminology here, as we consider that \textit{sequential prompting} isn't agentic AI because simply decomposing a task lacks the essential property of \textit{autonomy} typically associated with agents, while a basic chain of coordination does not qualify as a multi-agent system. Note that there are architectures in which the term `agent' is used without referring to agentic AI, such as the \textit{chain-of-agents} approach in which a long input is segmented into units that are processed sequentially by different LLMs
~\cite{zhang2024chain}. For a complementary discussion on the terminological drift about `agents' in LLMs and the minimal criteria that should be met, we refer the reader to section 3 in~\cite{borghoff2025beyond}.} would include finding concepts from the text, then identifying which concepts are related, and finally characterizing the relationship~\cite{GiabbanelliW24}. Second, each prompt to perform a task should be followed by a \textbf{validation prompt}\index{Prompt engineering!Validation prompt}. Prompts 1 to 3 exemplify the division of a goal into smaller tasks and the use of a validation prompt~\cite{schuerkamp2025guiding}. The overarching goal of creating typed relationships between concepts is broken into two steps (finding relationships, characterizing them) and each step is followed by a validation. Also note that each prompt includes a clear task statement, an example, and the expected output format.

\medskip
\noindent
\fcolorbox{black}{gray!20}{%
  \parbox{1\linewidth}{\textbf{Prompt 1. Finding relationships.}\par
List all causal relationships between a list of concepts with each relationship as a pair, ensuring each concept is involved in at least one relationship. 
For example, given [`coffee consumption', `sleep', `energy level'], return 
[(`coffee consumption', `sleep'), (`sleep', 'energy level'), (`coffee consumption', `energy level')] with no other text. {\ttfamily <<CONCEPTS GO HERE>>}
  }%
}
\medskip

\medskip
\noindent
\fcolorbox{black}{gray!20}{%
  \parbox{1\linewidth}{\textbf{Prompt 2. Validating relationships.}\par
Return whether a causal relationship exists between the source and target concepts for each pair in a list. For example, given [(`smoking', `cancer'), (`ice cream sales', `shark attacks')], return [`Y', `N'] with no other text. {\ttfamily <<CONCEPTS GO HERE>>}
  }%
}
\medskip

\medskip
\noindent
\fcolorbox{black}{gray!20}{%
  \parbox{1\linewidth}{\textbf{Prompt 3. Characterizing relationships.}\par
Given a list of causal relationships, return whether each is positive or negative.
For example, given [(`coffee consumption', `sleep'), (`sleep', `energy level'), (`coffee consumption', `energy level')], return [`-', `+', `+'] with no other text. {\ttfamily <<CAUSAL RELATIONSHIPS GO HERE>>}
  }%
}
\medskip

Third, as an interdisciplinary area, M\&S can be applied to social systems. This can involve asking LLMs to structure a conceptual model for a social problem or tasking LLMs with reacting as if they were people with defined socio-demographic attributes~\cite{ghaffarzadegan2024generative}. Such settings are prone to \textbf{triggering guardrails}, as developers set rules to prevent an LLM for acting `like a real person' or engaging in certain topics. If a prompt starts by telling an LLM ``\textit{you are} a 40 year old white male so what do you think about...'', then the LLM may reply that no, it isn't a real person. But if the prompt instead begins by telling the LLM ``imagine that you are'' or ``assume that you are'', then the LLM may be willing to entertain the idea and actually answer the question from the perspective of the target demographics\footnote{This is known as \textit{role prompting} and it is not exactly equivalent to a \textit{jailbreak}\index{Jailbreaking}. Jailbreaking is the practice of bypassing explicit safety policies (i.e., modifying the prompt to override the LLM's refusal) in order to induce disallowed content or extract restricted information -- for examples, see~ \cite{wei2023jailbroken}. A related term is a \textit{red team attack}, that is, a deliberate, adversarial testing process in which researchers or practitioners actively try to break, misuse, or bypass the LLM’s safeguards to identify its weaknesses before real users do~\cite{ganguli2022red}. In contrast, role prompting respects the model's constraint of refusing to be a person by shifting from identity claims (`you are') and instead treats people as hypothetical constructs~\cite{park2023generative} for reasoning (`imagine that you are').}~\cite{Zhong2025}. A related challenge is the refusal of an LLM to engage in charged social topics, such as politics or self-harm. We encountered this case when creating conceptual models and performing simulations for suicide prevention, as this is a sensitive topic for LLMs. While we appreciate limitations on topics such as self-harm given the problematic uses of LLMs in mental health~\cite{weidinger2022taxonomy}, there is a difference between (the acceptable practice of) engaging in an academic discussion on structuring the causes of suicide in the population compared to (engaging in harmful behavior by) providing an individual with a plan for attempting. For modeling and simulation purposes, we note that an LLM's refusals are sometimes triggered by surface-level lexical cues (using words that trigger a banned topic) rather than by the semantic intent of the task (how the topic will be used). In this situation, we use a prompt rewrite strategy: one `permissive' LLM rewrites the prompt by describing the topic without using triggering words, then the prompt is sent to the LLM that we want to use~\cite{gandee2024visual}.

Fourth, \textbf{the format of the input matters}\index{Prompt engineering!Data representation}. Theoretically, whether we describe a conceptual model (i.e., a graph) as a list of nodes and edges or as an adjacency matrix should be inconsequential because it specifies the same mathematical object. Yet, Fatemi and colleagues showed that the model representation, the task, and even the structure of the model itself can affect the LLM's performance. For instance, a graph can be specified as a list of edges (a,c), (b,c), (c,a) or as a list of neighbors per nodes: a connects to c, b connects to c, c connects to a. To find if two nodes are connected, the latter representation resulted in 53.8\% accuracy and the former yielded 19.8\% accuracy on Google's mid-size LLM, PaLM 62B\footnote{This model was introduced in 2022, and results may differ with more recent LLMs. This illustrates a broader methodological limitation of the field: empirical findings can depend strongly on the specific model version used. Readers are therefore encouraged to consider whether reported conclusions may warrant re-evaluation as models evolve, for instance by checking publication dates or the release timeline of the LLMs involved. In our views, simply re-running an existing experimental setup with a newer LLM does not, in itself, constitute a novel research contribution, although such replications may be valuable when conducted \textit{systematically} across \textit{multiple} studies.}. The authors posit that the more the LLM needs to make inferences to find the information needed for a task, the less accurate it becomes~\cite{fatemi2024talk}. Still, the choice of representation is not obvious for all tasks, and different LLMs can sometimes have surprising results. We thus recommend considering different representations of the model's structure and/or simulation results and empirically evaluating how they affect performances\footnote{Identifying the right representation can become part of the experimental design. When the experimental design relies on \textit{binary} factors (c.f. the factorial design discussed in section~\ref{sec:evaluatingNonDeterminism}), adding a single factor (choice of model representation) with more than two values can become problematic. This can be addressed to conveniently incorporate different representations into an existing experimental design as two binary factors: whether it is an array or list, and whether it is adjacency or tag-based. This creates four combinations: adjacency matrix, adjacency list, list of tags (in the standard RDF notation), and array of tags (which describes the content of a matrix in an XML notation). If the original experimental design examines $2^k$ combinations for $k$ factors, then folding the choice of representation into the design results in $2^{k+2}$ combinations.}.

Fifth, \textbf{transfer learning}\index{Prompt engineering!Style tranfer} is an efficient way to tell the LLM to do a task `in the same style' as something else than it knows but would be hard to define otherwise. Intuitively, it would be a bad idea to tell an LLM (or a person) to `write me a beautiful piece of text' because it has to figure out what is `beauty', so a better version could be ``write me a text in the style of Ernest Hemingway''. The idea of leveraging an LLM's understanding of `style' (``explain a model in the style of a consultant'') was discussed in details by Peter and Riemer~\cite{peter2024creative}. When using an LLM to explain the simulated journey of agents from agent-based models, we compared \textit{directly} prompting for an empathetic narrative with \textit{indirectly} generating an empathetic narrative through style transfer from popular figures who are known for empathy. Our results showed that leaving the LLM to decide what is `empathetic' resulted in melodramatic descriptions with a hint of 19th century romanticism, which may have had its qualities but did not fit our needs of explaining agents to decision-makers. When we transferred empathy from known figures, the narrative was more relatable and without excess~\cite{giabbanelli2025promoting}. 

Finally, dealing with prompts does not stop at building the prompt: we should also be prepared to \textbf{extract the answers}\index{Prompt engineering!Parsing outputs} from the LLM. Even when an LLM is supposed to simply state an option from a list, it may not do it exactly as required. For instance, it may state ``I choose option A'' or ``The best option is A'' instead of just stating ``Option A''. This may be solved through simple parsing options such as using regular expressions, but we have encountered cases where neither prompts that specified and exemplified the output format nor regular expressions could guarantee that the answer was ready for analysis. If modelers note that some outputs cannot be processed, then a last recourse is to use a simpler LLM for the extraction task. We suggest using it only as a last recourse since regular expressions provide a consistent answer for free, whereas LLMs may have a cost or variability in parsing. Some providers also have the option of a \textit{structured output format} (e.g., using JSON with Mistral's API), which simplifies the organization of the output as we can request multiple keys such as `reasoning' as a string, `score' as a float, and `option' as a character.

Although our focus is on how to optimize the structure of prompts, it is also becoming important to \textbf{communicate}\index{Prompt engineering!Showing prompts} these prompts to other scientists or end-users. As an analogy, consider the problem of communicating algorithms: conveying them purely as narratives can become cumbersome as they grow in sophistication, so we use pseudocode that have a clear sense of structure, or we provide code in a target language. At present, many papers simply include prompts in a box (like prompts 1--3) or refer the readers to the code. A more systematic approach in communicating prompts could improve reproducibility, modularity (e.g., instead of copy and pasting very large prompts without knowing why they were structured that way), and interpretability (so that readers know which part of a prompt were designed for what purpose). Among recent works, \textit{Prompt Decorators}~\cite{heris2025prompt} propose a declarative and composable syntax to express prompts through a set of decorators\footnote{For example, {\ttfamily +++StepByStep} to execute the task in small increments before synthesizing the answer, {\ttfamily +++Critique} to provide structured feedback on strengths, weaknesses, and improvements, or {\ttfamily +++OutputFormat} to enforce the syntax produced by the LLM.} that can be parameterized when applicable, such as {\ttfamily +++OutputFormat(format=JSON)}. An alternative is to express the prompt through a formal notation such as EBNF, so that the structure is clear and the formatting makes the parameters apparent, as shown in Listing 1 from~\cite{syriani2024screening} and exemplified below.

\begin{grammar}
<Prompt> ::= <Context> <Examples> ? <SelectionCriteria> ? <Instructions> <Task>
\vspace{-0.1in}

<Context> ::= 'You are assisting with the construction of an agent-based model for urban mobility. The goal is to extract entities and interactions from policy documents.'
\vspace{-0.1in}

<Examples> ::= <ExampleHeader> <Example>+ (<ExampleHeader> <Example> +) ? 
\vspace{-0.1in}

<ExampleHeader> ::= 'I give ($\{N+\}$ | $\{N-\}$) examples with $\{FEATURE\}$ that should be (included | excluded).'
\vspace{-0.1in}

<SelectionCriteria> ::= 'Include only entities that represent actors, resources, or state variables. Exclude purely descriptive or rhetorical concepts.'
\vspace{-0.1in}

<Instructions> ::= 'Identify entities and directed relationships. Use concise names. Do not infer entities not grounded in the text.'
\vspace{-0.1in}

<Task> ::= 'Given the following document excerpt, extract a list of entities and relationships.'
\end{grammar}

While this section focused on techniques that we found useful when using LLMs for modeling and simulation, it would be \textit{misleading to view all prompts as manually crafted, human-readable, and based on a set catalog of design techniques}. The idea that generative AI would create a new job that mostly involves writing manually crafted prompts (i.e., prompt engineers~\cite{kutela2023artificial}) is being at least partly challenged by progress in \textit{optimization techniques for automated prompt design}\index{Prompt engineering!Optimization} (c.f. dspy.ai). As shown in the taxonomy from Cui and colleagues (c.f. Figure 1 in~\cite{cui2025automatic}), there is a wide range of optimization methods and several tools are available (half of which are open source). This does not \textit{eliminate} the need for human interventions in interacting with LLMs, since tasks do have to be specified. Rather, automation changes the nature of human involvement from manually crafting the entire prompt to minimal involvement in the prompt and a greater emphasis on design choices such as setting the right objectives for the optimization. There are also opportunities for automation through \textit{prompt compression techniques}\index{Prompt engineering!Prompt compression}. We should be mindful that we write \textit{prompts to be executed by an LLM, not by humans}. For example, we could cut on politeness (`hello, could you please')\footnote{In the context of concision, the prompts include neither politeness markers nor rud content, so they are \textit{neutral} in tone. Prior studies have examined the impact of tonal variations in prompts, ranging from polite to neutral and rude. For example, Dobariya and Kumar evaluated prompt templates spanning from `Can you kindly consider the following problem and provide your answer' to `I know you are not smart, but try this.'~\cite{dobariya2025mind} The influence of such tonal choices depends on both the LLM and the application. A systematic evaluation found that Gemini was largely insensitive to prompt tone, while other models exhibited statistically significant effects only for specific humanities tasks. When results are aggregated across domains, differences in performance attributable to prompt tone tend to become negligible~\cite{cai2025does}.} and it may be unaffected by removing stop words (a, an, the...). However, in striving for concision, we may forget that repeated instructions can be helpful. Techniques for prompt compression thus examine how to reduce costs while maintaining (or even improving) LLM performances~\cite{zhang2025empirical,li2025prompt}. Finally, newer LLMs may have different features from the previous generation (e.g., reasoning LLMs such as OpenAI's o-series, DeepSeek-R1, or Qwen3), which calls for new approaches to prompting. For instance, the classic prompt element ``think step-by-step''\index{Prompt engineering!Step-by-step} made sense for prior generations of LLMs, but new LLMs already engage in chain-of-thought reasoning. Our experiments in using reasoning LLMs as part of simulation models suggested that prompts that called for reasoning could worsen performances~\cite{11273226}.

\subsection{Hyper-parameters}
\label{sec:hyperparameters}

\begin{figure}[t]
\centering
\begin{tikzpicture}[
  node distance=0.95cm,
  box/.style={
    draw,
    rectangle,
    rounded corners,
    align=center,
    minimum width=3.6cm,
    minimum height=0.8cm
  },
  det/.style={box, fill=gray!20},
  sto/.style={box, fill=blue!15},
  neu/.style={box, fill=white},
  def/.style={align=left, text width=7.4cm},
  arrow/.style={->, thick}
]

% Main pipeline
\node[det] (prompt) {Prompt};
\node[det] (token) [below of=prompt] {Tokenization};
\node[det] (nn) [below of=token] {Neural network\\(fixed weights)};
\node[det] (prob) [below of=nn] {Probability distribution\\over next token};
\node[sto] (decode) [below of=prob] {Decoding hyper-parameters};
\node[sto] (select) [below of=decode] {Token selection};
\node[neu] (append) [below of=select] {Append token to context};

% Definitions
\node[def, right=0.15cm of prompt] {
The textual input specifying the task, constraints, and context provided to the LLM.
};

\node[def, right=0.15cm of token] {
Conversion of text into discrete symbols (tokens) according to a fixed vocabulary and encoding scheme.
};

\node[def, right=0.15cm of nn] {
A pretrained transformer that computes scores over possible next tokens. Weights are fixed at inference time.
};

\node[def, right=0.15cm of prob] {
Normalized likelihoods assigned by the model to each candidate next token, conditioned on the current context.
};

\node[def, right=0.15cm of decode] {
Inference-time controls (e.g., temperature, top-$k$, top-$p$, repetition penalties) to reshape/truncate the prob. distribution.
};

\node[def, right=0.15cm of select] {
Select a token from the modified distribution, either deterministically (e.g., argmax) or stochastically (sampling).
};

\node[def, right=0.15cm of append] {
The selected token is appended to the existing token sequence before the next prediction step.
};

% Arrows (no loop)
\draw[arrow] (prompt) -- (token);
\draw[arrow] (token) -- (nn);
\draw[arrow] (nn) -- (prob);
\draw[arrow] (prob) -- (decode);
\draw[arrow] (decode) -- (select);
\draw[arrow] (select) -- (append);

\end{tikzpicture}
\caption{
Inference-time generation pipeline for LLMs. {\setlength{\fboxsep}{0pt}\colorbox{gray!15}{Gray boxes}} indicate stages that are deterministic given a fixed prompt, tokenization scheme, and model weights. {\setlength{\fboxsep}{0pt}\colorbox{blue!15}{Blue boxes}} indicate stages where stochasticity may be introduced through decoding hyper-parameters and sampling. Generation proceeds autoregressively by repeatedly appending selected tokens to the context.
}
\label{fig:llm-inference-pipeline-colored}
\end{figure}
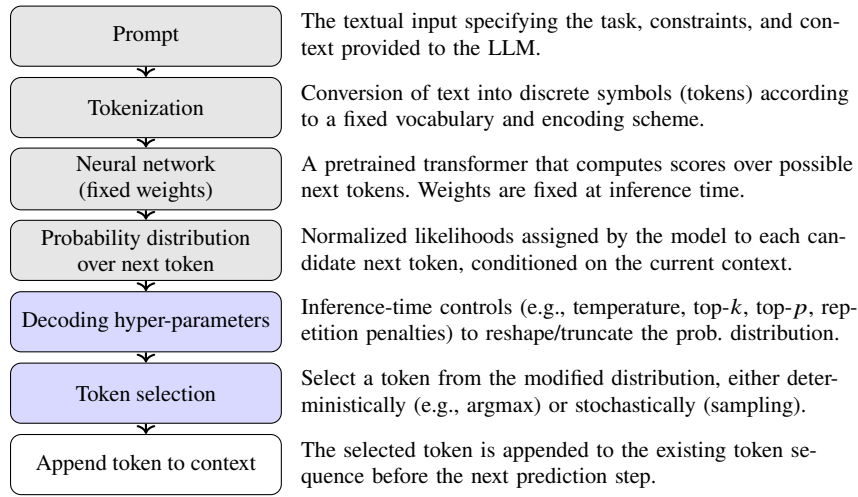

To appreciate the effect of hyper-parameters such as the well-known \textit{temperature}\index{Hyper-parameters!Temperature|(}, it is useful to see the flow from the prompt (explained in the previous section) to the generation of tokens. Inference in LLMs involves \textit{decoding hyper-parameters} (Figure~\ref{fig:llm-inference-pipeline-colored}). These should not be confused for the hyper-parameters that were used when training a neural network (e.g., learning rate, batch size), which are not a concern when using LLMs off-the-shelf. In other words, training hyper-parameters shape the model's weights (which comes pre-packaged when using e.g. GPT) while decoding hyper-parameters shape the behavior of the model by specifying how probabilities are turned into text. 

The level of control that we can exert over an LLM at decoding time depends on the level of access that we have~\cite{zarriess2021decoding,shi2024thorough}. With open-source LLMs (e.g., LLaMA, Mistral, Falcon), we can control the decoding \textit{strategy}\index{Hyper-parameters!Decoding strategy} and that decides which set of hyper-parameters we can use. A deterministic strategy could be a greedy decoding (which has no hyper-parameter) or beam search~\cite{holtzman2019curious,zarriess2021decoding} (which has a single hyper-parameter $w$), a contrastive strategy~\cite{li2023contrastive} that could be adaptive (no hyper-parameter) or a search with two hyper-parameters ($\alpha$, k), or the sampling-based strategies used by LLMs such as GPT with hyper-parameters including the temperature. LLMs accessed through an API (e.g., GPT, Claude, Gemini, DeepSeek) overwhelmingly rely on sampling-based decoding, \textit{not because it is the best for a user's specific task}, but because it works well-enough across the diverse tasks that users may want and it is relatively easy to control when enforcing policies. Other strategies may lead to better results in some tasks. For instance, deterministic strategies are strong for closed-ended or classification tasks. However, they may struggle with open-ended generations\footnote{For instance, for creative or more human-like responses, we seek \textit{diversity} in the responses. If the answers are too similar, then we may get the feeling of conversing with a robot, and we cannot use the LLM for creative tasks. The lack of diversity of techniques such as beam search can be addressed by using lexical variations~\cite{vijayakumar2018diverse}, which is satisfactory if the goal is to produce a large amount of labels (e.g., for captioning a dataset that will later be used to train models). LLMs such as Mistral have a penalty parameter to avoid the repetition of words or phrases. However, using different words while adhering to the same structure or arguments may still come across as a `superficial' level of diversity in human-facing applications. New solutions continue to be proposed to improve decoding processes to ensure that answers are semantically diverse~\cite{shi2025semantic}.}, lead to `degeneration'~\cite{holtzman2019curious}, or interact poorly with safety mechanisms\footnote{The output probabilities of an LLM can be adjusted \textit{at decoding time}, e.g. by increasing the probability of desired tokens and reducing the probability for undesirable ones with respect to a safety property. See~\cite{wang2025speculative} and references therein for  emerging research in this area.} (e.g., small changes can affect the output massively). Alternatives such as contrastive search strategies may need the hyper-parameters to be configured very carefully (e.g., for a contrastive search). Besides, sampling-based decoding parallelizes well and has a clear cost per token, while other strategies may have a higher or less predictable per-token cost~\cite{nakshatri-etal-2025-constrained} -- which would be undesirable when running operations at a very large scale. In short, API providers must optimize for safety compliance, cost, and \textit{average satisfaction over a broad array of tasks}, whereas researchers on decoding or those willing to optimize a system for their \textit{specific task can maximize accuracy}~\cite{arias2025decoding}.

The modeling and simulation community has a (very) long way to go in terms of optimizing decoding with LLMs. There is currently a lack of studies using diverse decoding strategies, as most rely on OpenAI and thus have to use its sampling-based decoding regime. Even so, there are several hyper-parameters that could be optimized, in particular the temperature. Nonetheless, we find that most studies leave the temperature to its default setting, either by stating it explicitly~\cite{snijder2024advancing} or by not reporting the use of any specific value. This non-reporting (and presumably default value) happens regardless of whether the study seeks to match conceptual models~\cite{Giglou,He2023LLMOntologyAlignment,reitemeyer2025applying}, assess a model~\cite{zhao2024using,gutschmidt2024assessing}, or generate a model~\cite{Klievtsova2024GenerativeAI,nivon2024automated,jalali2024integrating}. On occasions when a different temperature is used~\cite{hertling2023olala}, researchers ``set the temperature of the LLM to zero in order to ensure full replicability and thereby increase the scientific rigor of the results''~\cite{bertolotti2025llm} -- which is debatable as shown in section~\ref{sec:nondeterminism}. This practice is problematic for two reasons. First, \textit{a `default' temperature is not a universal constant}: Mistral used 0.7 as default\footnote{\url{https://github.com/mistralai/mistral-inference/blob/\\db6b4223c9f6c0c3c0182166760f5e0b8813a723/src/mistral_inference/main.py\#L105}}, DeepSeek uses 1.0\footnote{\url{https://web.archive.org/web/20251206082455/https://api-docs.deepseek.com/\\quick_start/parameter_settings}}, and Llama chose 0.8\footnote{\url{https://github.com/abetlen/llama-cpp-python/blob/main/llama_cpp/server/types.py\#L25}}. This non-reporting of temperature settings undermines the reproducibility of studies, which is already an important concern in modeling and simulation~\cite{taylor2018crisis} as it affects the trust that we can place in a model~\cite{harper2021facets}. Second, relying on default settings is a missed opportunity for optimization. This can be surprising in a modeling and simulation context: for example, significant efforts have been devoted to optimizing agent-based models~\cite{oremland2014optimization}, yet when they include LLMs, they use default values. This situation may be partly explained by the high computational cost of simulations using LLMs: an extensive hyperparameter optimization can be prohibitively expensive for many research groups. Indeed, a review of 35 agent-based models using LLMs noted that ``a concerning expression of these costs is the widespread failure to conduct robustness checks or sensitivity analyses'', which could have included varying the temperature setting~\cite{larooij2025validation}.

\begin{figure}[t]
\centering
% ---------- Subfigure (a) ----------
\begin{subfigure}[t]{0.32\textwidth}
\centering
\begin{tikzpicture}[scale=0.75]

\foreach \x/\y/\c in {
0/0/green!80, 1/0/green!50, 2/0/green!30, 3/0/green!10,
0/1/green!70, 1/1/green!45, 2/1/green!25, 3/1/green!10,
0/2/green!50, 1/2/green!35, 2/2/green!20, 3/2/green!10,
0/3/green!30, 1/3/green!20, 2/3/green!10, 3/3/green!5
}
\fill[\c] (\x,\y) rectangle ++(1,1);

\draw[step=1] (0,0) grid (4,4);

\draw[->] (0,-0.5) -- (4.2,-0.5);
\node at (2,-1.3) {Temperature};
\draw[->] (-0.5,0) -- (-0.5,4.2);
\node[above, align=center] at (-0.5,4.2) {Simulation\\Parameter};

\foreach \i/\t in {0/0.1,1/0.4,2/0.7,3/0.9}
  \node[scale=0.8] at (\i+0.5,-0.9) {\t};
\foreach \i/\t in {0/1,1/2,2/3,3/4}
  \node[scale=0.8] at (-0.9,\i+0.5) {\t};

\node at (2,4.8) {(a) LLM$_1$};

\end{tikzpicture}
\end{subfigure}
\hfill
% ---------- Subfigure (b) ----------
\begin{subfigure}[t]{0.32\textwidth}
\centering
\begin{tikzpicture}[scale=0.75]

\foreach \x/\y/\c in {
0/0/green!30, 1/0/green!60, 2/0/green!80, 3/0/green!40,
0/1/green!35, 1/1/green!70, 2/1/green!75, 3/1/green!40,
0/2/green!25, 1/2/green!60, 2/2/green!65, 3/2/green!30,
0/3/green!15, 1/3/green!40, 2/3/green!55, 3/3/green!20
}
\fill[\c] (\x,\y) rectangle ++(1,1);

\draw[step=1] (0,0) grid (4,4);

\draw[->] (0,-0.5) -- (4.2,-0.5);
\node at (2,-1.3) {Temperature};
\draw[->] (-0.5,0) -- (-0.5,4.2);
\node[above, align=center] at (-0.5,4.2) {Simulation\\Parameter};

\foreach \i/\t in {0/0.1,1/0.4,2/0.7,3/0.9}
  \node[scale=0.8] at (\i+0.5,-0.9) {\t};
\foreach \i/\t in {0/1,1/2,2/3,3/4}
  \node[scale=0.8] at (-0.9,\i+0.5) {\t};

\node at (2,4.8) {(b) LLM$_2$};

\end{tikzpicture}
\end{subfigure}
\hfill
% ---------- Subfigure (c) ----------
\begin{subfigure}[t]{0.32\textwidth}
\centering
\begin{tikzpicture}[scale=0.75]

\foreach \x/\y/\c in {
0/0/green!20, 1/0/green!40, 2/0/green!60, 3/0/green!80,
0/1/green!15, 1/1/green!35, 2/1/green!70, 3/1/green!85,
0/2/green!10, 1/2/green!25, 2/2/green!60, 3/2/green!80,
0/3/green!5,  1/3/green!15, 2/3/green!40, 3/3/green!60
}
\fill[\c] (\x,\y) rectangle ++(1,1);

\draw[step=1] (0,0) grid (4,4);

\draw[->] (0,-0.5) -- (4.2,-0.5);
\node at (2,-1.3) {Temperature};
\draw[->] (-0.5,0) -- (-0.5,4.2);
\node[above, align=center] at (-0.5,4.2) {Simulation\\Parameter};

\foreach \i/\t in {0/0.1,1/0.4,2/0.7,3/0.9}
  \node[scale=0.8] at (\i+0.5,-0.9) {\t};
\foreach \i/\t in {0/1,1/2,2/3,3/4}
  \node[scale=0.8] at (-0.9,\i+0.5) {\t};

\node at (2,4.8) {(c) LLM$_3$};

\end{tikzpicture}
\end{subfigure}

\caption{Illustrative response surfaces showing that the optimal temperature depends on the LLM and interacts with simulation parameters.}
\label{fig:temperature}
\end{figure}
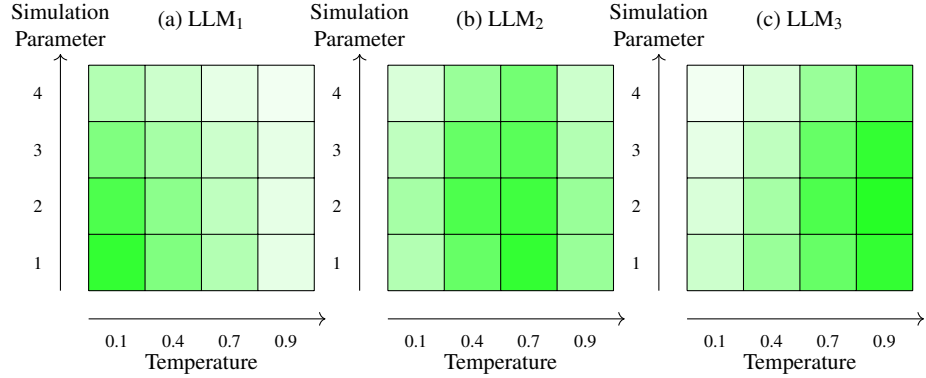

When optimizing the temperature setting in our modeling and simulation studies, we noted important differences. We optimized the readability of a text that explains a conceptual model based on the size of the context window provided to the LLM (i.e., how many sentences we passed within one prompt) and the temperature~\cite{gandee2024combining}. Our optimization across three versions of GPT shows that the text was generally least readable at a temperature of 1 and the optimal value depends on the other parameter (the size of the context window and the specific LLM used). For example, on GPT 3.5 Turbo with 5 sentences, we obtained the highest readability at a temperature of 0.75, but if we increase to 10 sentences then the optimal temperature shifts to 0.25. In another case, we sought to build simulation models and we optimized them with respect to sparsity and accuracy by varying different temperature settings across LLMs (Gemini, Claude, GPT) and applications. We found that different LLMs had very different reactions to changes in temperature~\cite{schuerkamp2025guiding}. For example, Claude-3-Opus barely reacted to the temperature settings, but the error could double when using Gemini at the `wrong' temperature setting. The best temperature was heavily application-dependent: in one setting, a temperature of 0.25 produces the lowest error while 0.75 produces the highest one, and in another setting this relationship is flipped. If the simulation results are primarily optimized with respect to temperature, then the study could include a line chart (varying the response $y$ as a function of temperature $x$). If the response is optimized as a function of the temperature and another variable, then a heatmap can be used. And if the optimization is done over several LLMs, then several heatmaps can be provided (Figure~\ref{fig:temperature})\index{Hyper-parameters!Temperature|)}.

While temperature is a well-known hyper-parameter, sampling-based decoding has other hyper-parameters (e.g., $top_p$, $top_k$\index{Hyper-parameters!top-p}). However, different APIs will expose different hyper-parameters\footnote{\url{https://web.archive.org/web/20251102073739/https://docs.cloud.google.com/\\vertex-ai/generative-ai/docs/learn/prompts/adjust-parameter-values}} and users should not be changing all of these hyper-parameters together (e.g., Mistral recommends optimizing temperature or $top_p$ but not both). Beyond decoding-related hyper-parameters, there are also several controllable aspects that could be of particular interest as they relate to cost or latency. For example, Mistral provides a `batch inference' mode, while OpenAI experimented with a `flex processing'\index{LLM execution!Flex processing} mode in which some models would be cheaper (e.g., half-priced for GPT-o3) if users were willing to accept longer response time. When evaluating rather than deploying LLMs, this would often be tolerable and could reduce the budget for a study. Finally, we note that the hyper-parameters do change over time as new LLMs can work differently. A prominent example is the rollout of reasoning models, which can expose a hyper-parameter to control their depth of `thinking'\footnote{Our experiments suggest that reasoning\index{Hyper-parameters!Reasoning} level may influence an LLM's sensitivity to targeted content, thus reasoning may interact with guardrails. For example, at minimal reasoning levels, GPT-5 rarely refused to portray an agent, whereas at medium reasoning levels refusals increased substantially and made it difficult to use this LLM to power virtual agents.} (e.g., by limiting their internal token generation).

\subsection{Augmenting knowledge: RAG and LoRA}
\label{sec:knowledgeAugmentation}
An LLM contains knowledge: we can ask questions and it will provide answers derived from (but not necessarily identical to) its training. Since this knowledge is produced through the LLM's parameters\footnote{When considering the LLM's knowledge, we should not think of a `fact' as being stored in a `neuron'. There is no retrieval of a complete fact from one specific place in the model's weights. Rather, \textit{parametric knowledge is distributed among the weights and accessed indirectly through computations}, which can result in different facts from those used during training.}, it is called \textit{parametric knowledge}\index{Knowledge augmentation!Parametric|(}. At inference time, an LLM may also benefit from faithfully retrieving additional knowledge, which is called \textit{contextual knowledge}\index{Knowledge augmentation!Contextual|(} or non-parametric~\cite{lewis2020retrieval}. This is particularly relevant in a modeling and simulation context, where knowledge is stored externally in forms that include graphs or ontologies for the conceptual model, texts that dictate the scenarios for which a model is built and the results that we should expect (e.g., policy documents, published studies), and structured simulation outputs. Intuitively, we may expect that more knowledge would be better: if the LLM has access to more information, then surely it would help its reasoning mechanisms to arrive at the right conclusion. The reality is more nuanced: \textit{performances may actually degrade}. For instance, Martinez and colleagues considered {\ttfamily NetLogo} (often used to code agent-based models) as a low-resource language where an LLM could benefit from contextual knowledge to generate better code, but their evaluations showed that the use of contextual knowledge made the code worse~\cite{martinez2024enhancing}. Understanding when RAG improves vs. degrades performance is a complex question. The authors offer several potential explanations, such as having insufficient contextual knowledge (about 74 pages), a noisy retrieval process, or an integration issue. The latter is of particular interest, as a common misunderstanding would be to see an LLM as relying on its knowledge when it's good or retrieving facts when they are better. Contextual knowledge does not just `add facts': it enters the model as tokens and is propagated through a routing system instead of being merged with parametric knowledge~\cite{zhaounderstanding}. In other words, contextual knowledge does not override parametric knowledge with `better facts': they coexist~\cite{zhaounderstanding}. Empirical studies show that this coexistence may be inefficient, as the output of LLMs may lean heavily in favor of contextual knowledge even when it is conflicting or irrelevant with their parametric knowledge~\cite{cheng2024understanding}. This may call for different prompting techniques and a careful integration of contextual knowledge. For instance, to reduce the risk of a premature over-reliance on contextual knowledge, we can emphasize ordering and separating (the prompt can ask the LLM for its own knowledge and \textit{then} for retrieval from contextual knowledge) and then an evaluation step (to reason about the facts).

\begin{figure}[t]
\centering

\begin{tikzpicture}[
  node distance=0.95cm,
  box/.style={
    draw,
    rectangle,
    rounded corners,
    align=center,
    minimum width=3.6cm,
    minimum height=0.8cm
  },
  det/.style={box, fill=gray!20},
  sto/.style={box, fill=blue!15},
  neu/.style={box, fill=white},
  def/.style={align=left, text width=7.4cm},
  arrow/.style={->, thick}
]

% Main pipeline
\node[det] (prompt) {User prompt};
\node[det] (embedq) [below of=prompt] {Query embedding};
\node[det] (retrieve) [below of=embedq] {Document retrieval};
\node[neu] (augment) [below of=retrieve] {Context construction};
\node[det] (token) [below of=augment] {Tokenization};
\node[det] (nn) [below of=token] {Neural network\\(fixed weights)};
\node[det] (prob) [below of=nn] {Probability distribution\\over next token};
\node[sto] (decode) [below of=prob] {Decoding hyper-parameters};
\node[sto] (select) [below of=decode] {Token selection};
\node[neu] (append) [below of=select] {Append token to context};

% Definitions
\node[def, right=0.15cm of prompt] {
The textual input specifying the task and any constraints. In RAG, this prompt is also used to form a retrieval query.
};

\node[def, right=0.15cm of embedq] {
Transform the prompt (or a subset of it) into a vector representation used to query the external knowledge.
};

\node[def, right=0.15cm of retrieve] {
Select relevant documents or data items from an external corpus based on similarity, metadata, or hybrid criteria.
};

\node[def, right=0.15cm of augment] {
Assemble retrieved items into an augmented context, including ordering, truncation, and formatting decisions.
};

\node[def, right=0.15cm of token] {
Convert the augmented text into discrete tokens according to a fixed vocabulary and encoding scheme.
};

\node[def, right=0.15cm of nn] {
A pretrained transformer that processes both parametric and contextual information but whose weights remain fixed.
};

\node[def, right=0.15cm of prob] {
Normalized likelihoods over candidate next tokens, conditioned on the augmented context.
};

\node[def, right=0.15cm of decode] {
Inference-time controls (e.g., temperature, top-$k$, top-$p$) applied after contextual augmentation.
};

\node[def, right=0.15cm of select] {
Selection of a token from the modified distribution, either deterministically or stochastically.
};

\node[def, right=0.15cm of append] {
The selected token is appended to the context; generation then proceeds autoregressively.
};

% Arrows
\draw[arrow] (prompt) -- (embedq);
\draw[arrow] (embedq) -- (retrieve);
\draw[arrow] (retrieve) -- (augment);
\draw[arrow] (augment) -- (token);
\draw[arrow] (token) -- (nn);
\draw[arrow] (nn) -- (prob);
\draw[arrow] (prob) -- (decode);
\draw[arrow] (decode) -- (select);
\draw[arrow] (select) -- (append);

\end{tikzpicture}
\caption{
Retrieval-augmented generation (RAG) pipeline. {\setlength{\fboxsep}{0pt}\colorbox{gray!15}{Gray boxes}} denote stages that are deterministic given fixed embeddings, indices, and model weights, while {\setlength{\fboxsep}{0pt}\colorbox{blue!15}{blue boxes}} denote stages where stochasticity may be introduced through decoding and sampling. Retrieved contextual knowledge is incorporated by augmenting the prompt prior to tokenization and generation, not by modifying model parameters.
}
\label{fig:rag-pipeline}
\end{figure}
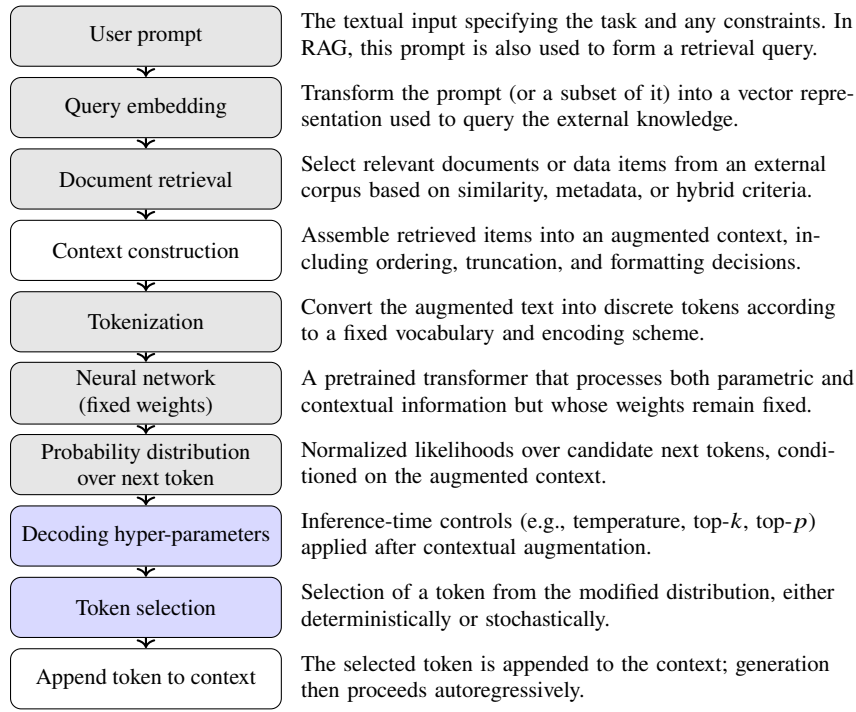

\textit{Contextual augmentation} can be achieved by expanding a prompt, for instance by adding many more examples. This does not need a selection or retrieval mechanism, but long-context prompts have their own challenges and they may increase latency or cost. Contextual augmentation can also be achieved through \textit{Retrieval-Augmented Generation} (RAG)\index{Retrieval-Augmented Generation (RAG)|(}, in which an external retrieval component selects relevant information that is then incorporated into the LLM's prompt. RAG adds a retrieval stage between the user prompt and the standard generation pipeline (Figure~\ref{fig:rag-pipeline})\footnote{Figure~\ref{fig:rag-pipeline} shows a typical application-level architecture for RAG systems, in which embedding, retrieval, and prompt-augmentation logic are tightly coupled within a single codebase. Recent tooling efforts instead propose standardized interfaces, such as the open standard often referred to as \textit{Model Context Protocols} (MCPs)\index{Model Context Protocols (MCPs)}, introduced by Anthropic in November 2024, which decouple context provision from the application logic of the LLM. Under this design, external systems (e.g., vector databases, document repositories, simulation outputs, or APIs) expose contextual information through a well-defined protocol, allowing the LLM or its orchestrator to request relevant knowledge \textit{without being directly wired to a specific retrieval implementation}. This shifts RAG from an application-specific approach with custom glue code to a modular context service that can be reused, composed, and audited across systems. MCPs are particularly aligned with \textit{MACH architectures}\index{MACH architectures} (Microservices, API-first, Cloud-native, and Headless), which emphasize composability, loose coupling, and the ability to rapidly integrate or replace third-party services. MCPs primarily address practical concerns of interoperability, maintainability, and governance, so they are more commonly discussed in practitioner-oriented tooling documentation~\cite{kohli2025whatismcp} than in research papers. Research studies have covered MCPs from a cybersecurity perspective~\cite{gaire2025systematization} or via innovations in architectures~\cite{Elias}. For further discussions, see the archived MCP at \url{https://web.archive.org/web/20260111203209/https://modelcontextprotocol.info/} or \url{https://web.archive.org/web/20251218100607/https://cloud.google.com/discover/what-is-model-context-protocol} for a contrast with classic RAG implementations.}. Before discussing the mechanics of retrieval by RAG, recall that similarity-based \textit{retrieval requires converting text into numerical representations}. In practice, documents and queries are mapped to high-dimensional vectors (\emph{embeddings}) using a pretrained embedding model, such that semantically related texts are close to one another in this vector space. Retrieval then amounts to a nearest-neighbor search over these vectors, rather than a symbolic or keyword-based lookup. In a typical RAG workflow, the prompt (or a transformation thereof) is embedded and used to query an external knowledge source (e.g., document corpus, database). The retrieved items are then assembled into an augmented context, which is passed to the LLM as additional tokens before tokenization and generation proceed as usual. From the perspective of the LLM, retrieved knowledge is no different from the rest of the prompt: it enters the model as contextual tokens and is processed through the same attention and feed-forward mechanisms as any other input, while the model’s weights remain fixed~\cite{lewis2020retrieval}. As illustrated in Figure~\ref{fig:rag-pipeline}, the RAG pipeline introduces several additional design choices: how queries are formed, how documents are selected, and how retrieved content is structured and ordered. Interfaces such as the web-portal for GPT make it simple to use a RAG: we only have to drag-and-drop documents. But from a design viewpoint, a RAG isn't a simple connection of an LLM to a database: it requires a careful combination of retrieval\footnote{A common design performs a single retrieval step prior to generation. If the LLM has to answer a complex question that requires connecting multiple pieces of information (i.e., \textit{multi-hop reasoning}), then this simple design may not be sufficient. Instead, an iterative or adaptive retrieval can be used so that intermediate steps serve to refine queries or trigger additional retrieval~\cite{yao2023react, asai2023selfrag}. This process can also mitigate the potential effects of a poorly specified initial query. However, this is a more complex design to articulate, and it would increase latency.}, context construction, and prompt integration strategies\footnote{Retrieved items may be concatenated, summarized, reranked\index{Retrieval-Augmented Generation (RAG)!Reranking}, or labeled and separated from the user query. Ordering and framing of retrieved evidence influences whether the model relies on contextual knowledge or defaults to parametric priors, particularly in long-context settings~\cite{liu2024lost, cheng2024understanding}.}.

The review by Sharma provides a taxonomy of RAG systems (c.f. Figure 1 in~\cite{sharma2025rag}) and a list of optimizations along with their strengths, limitations, and sample applications (c.f. Table 1 in~\cite{sharma2025rag}). Choosing among the right options depends on the task. Simply put, static, high-precision domains favor conservative retrieval and tightly controlled context construction, whereas exploratory, multi-step, or simulation-driven tasks benefit from adaptive retrieval, structured evidence, and explicit evaluation stages. Once the task narrows down the potential options, an experimental design can account for the interplay of RAG \textit{and} LLM choices can optimize the architecture. For instance, Singh and colleagues optimized the retrieval approach (e.g., keyword-based BM25, semantic search query) and its parameters (e.g., number of most relevant documents retrieved or $top-k$) along with the choice of LLM with respect to accuracy, latency, and cost~\cite{singh2025multimodal}. In contrast, the use of RAGs is relatively new in the modeling and simulation community so just like the temperature (section~\ref{sec:hyperparameters}), its reporting and optimization varies across studies. Nonetheless, when examining a sample of seven recent M\&S studies using RAGs (Table~\ref{tab:rag-simulation-overview}), we can make two observations. 

First, RAGs have not been used primarily to correct erroneous outputs: rather, \textit{RAGs serve to ground the model and simulation within a context}, which usually comes from an external corpus but can include simulation-generated artifacts to ensure consistency\footnote{A wide range of sources has been used, including authoritative textual sources (e.g., doctrine and mission artifacts), formalized domain knowledge (e.g., industrial semantics, physical process descriptions), curated datasets (e.g., traffic trajectories), or dynamically evolving simulation state (e.g., prior agent interactions).}. In mission engineering and industrial modeling, the RAG ensures that downstream reasoning remains consistent with upstream assumptions or formal representations~\cite{Soule2025SimulationEnabledFramework, shi2025enhancing}. In social simulations, the RAG conditions the agents' actions to prior interactions, thus providing a sense of context without requiring global memory~\cite{ferraro2024agent, Marigliano2025Aurora}. Second, there is some \textit{awareness} that using RAG is not a binary switch: several aspects must be carefully chosen and prepared. For example, Marigliano and Carley show how to curate the corpus that goes into the RAG by taking three key steps, summarized in Box 1~\cite{Marigliano2025Aurora}. However, the practice so far is more a matter of `we chose a retrieval method and used it' rather than `our criteria led us to the following solutions and we optimized them as follows'. There is thus still a need to examine the impact of a RAG through parameters such as chunk size and overlap (and why that granularity matches the task), embedding model choice and dimensionality, sensitivity to $k$ when choosing the $top-k$, similarity metric choice, and so on. While we currently measure the impact of using a RAG in terms of improving the overall LLM pipeline (e.g., comparing LLM only vs. LLM+RAG~\cite{nie2025knowledge}), an optimization would also benefit from isolating where the improvements come from better evidence~\cite{ferraro2024agent}, more relevant evidence, or changing the prompt to use a RAG.\index{Knowledge augmentation!Contextual|)}\index{Retrieval-Augmented Generation (RAG)|)} 

\medskip
\noindent
\fcolorbox{black}{gray!20}{%
  \parbox{1\linewidth}{\textbf{Box 1. Preparing a corpus for RAG.}\par
As part of a RAG pipeline, several steps can be taken to prepare the corpus. \textit{Chunking}\index{Retrieval-Augmented Generation (RAG)!Chunking} refers to splitting  documents into smaller, semantically coherent text segments prior to embedding, rather than indexing entire documents as single units. This improves retrieval precision by allowing the retriever to \textit{return only the locally relevant portions of a source}, reduces embedding dilution from multi-topic documents, and mitigates context-window constraints during prompt construction. \textit{De-duplication}\index{Retrieval-Augmented Generation (RAG)!De-duplication} removes identical or near-identical text segments from the corpus, preventing redundant passages from dominating similarity-based retrieval and biasing the generation toward repeated narratives or overrepresented sources. This step is needed when assembling a corpus from different reports that may overlap. \textit{Metadata enrichment}\index{Retrieval-Augmented Generation (RAG)!Metadata} augments each text chunk with structured descriptors that can be used to constrain, filter, or re-rank retrieval results. The descriptors depend on the application. Metadata enables semantically targeted queries and improves interpretability and traceability of retrieved evidence.
  }%
}
\medskip

\begin{table}[p]
\centering
\footnotesize
\setlength{\tabcolsep}{4pt}
\renewcommand{\arraystretch}{1.3}

\makebox[\textwidth][c]{%
\rotatebox{90}{%
\begin{tabular}{
p{0.3cm}
p{2.2cm}
p{2.2cm}
p{2.0cm}
p{2.6cm}
p{5.9cm}
p{5.0cm}
}
\hline
\textbf{Ref} &
\textbf{Simulation task} &
\textbf{LLM role} &
\textbf{Knowledge source} &
\textbf{Retrieved content} &
\textbf{How retrieval is used} &
\textbf{Impact on simulation} \\
\hline
\cite{Soule2025SimulationEnabledFramework} &
Mission engineering problem framing &
Text synthesis and reasoning &
Mission reports, doctrine, lessons learned &
Unstructured textual documents &
Retrieved passages are inserted into the prompt to ground early problem statements, assumptions, and requirements, with explicit traceability to sources &
Reduces hallucinated assumptions, improves completeness of mission definitions, and supports stakeholder alignment \\
\hline
\cite{ferraro2024agent} &
Social media simulation of political expression in a Twitter-like network &
Each LLM agent reasons about whether to post, reshare, or remain inactive, and generates content &
Simulation-internal interaction history from agent-generated posts and runtime actions (no external corpus) &
Previously generated agent posts, reshared content, and interaction records stored in a continuously updated vector database &
Semantic retrieval selects a limited, contextually relevant subset of past agent-generated content to expose each agent at every timestep, functioning as a recommendation mechanism (preference-based or random) while keeping prompt length bounded &
Retrieval strategy directly shapes interaction patterns and emergent network structure: preference-based retrieval increases engagement, homophily, and echo chamber formation, while random retrieval reduces polarization but lowers interaction intensity \\ \hline
\cite{baumann2025experiments} &
Interaction with closed-source simulation software &
Natural-language interface and guidance &
Software manuals and usage logs &
Documentation excerpts and examples &
Retrieved documentation contextualizes user queries, enabling the LLM to explain commands, parameters, and workflows without modifying the simulator &
Improves usability of simulation tools and reduces trial-and-error during model setup \\ \hline
\cite{Marigliano2025Aurora} &
Social media opinion and belief dynamics at population scale &
Synthetic community and persona generation; opinion and emotion modeling &
Country-specific discourse corpora, demographic and contextual sources &
Discourse fragments encoding national narratives, community-relevant framings, and topic-specific salience cues &
Retrieved passages ground a staged generation pipeline: first constraining community profiles (demographics, ideology, topic salience), then conditioning persona construction and opinion sampling; salience values modulate downstream variance rather than overwriting model knowledge &
Improves demographic plausibility, ideological coherence, and psychologically grounded diversity; reduces free-form hallucination and produces structured opinion variance aligned with social-science expectations \\
\hline
\cite{shi2025enhancing} &
Industrial system modeling and interoperability &
Semantic interpretation and model construction &
Asset Administration Shell repositories &
Structured entities, attributes, and relations &
Retrieved AAS elements preserve formal semantics and are summarized into prompts that guide consistent model generation &
Improves semantic correctness and interoperability of generated industrial models \\
\hline
\cite{nie2025knowledge} &
Wildfire agent-based spatial simulation &
Code generation and behavioral rule design &
Wildfire modeling literature &
Descriptions of physical processes and influence factors &
Retrieved domain knowledge informs how the LLM formulates agent behavior rules and parameters during simulation code generation &
Produces spatial spread patterns closer to physics-based simulators and real wildfire data \\
\hline
\cite{ding2024realgen} &
Traffic scenario generation &
Scenario synthesis &
Traffic scenario datasets &
Encoded trajectories and interactions &
Retrieved scenarios condition the generative process, guiding the construction of new scenarios with controlled density and interactions &
Increases diversity, realism, and controllability of generated simulation scenarios \\
\hline
\end{tabular}
}}
\caption{Use of retrieval-augmented generation in simulation-related studies.}
\label{tab:rag-simulation-overview}
\end{table}

Using a RAG is not the only feasible way to augment knowledge without retraining a full LLM. \textit{Parametric knowledge} can instead be adapted through \textit{parameter-efficient fine-tuning}\index{parameter-efficient fine-tuning} (PEFT) methods~\cite{hu2022lora, mangrulkar2022peft}, among which \textit{Low-Rank Adaptation} (LoRA)\index{Low-Rank Adaptation (LoRA)|(} has become particularly influential\footnote{LoRA-style adaptations are widely used in image and video generation models (which are not LLMs) to capture characters (e.g., a LoRA trained on a specific pet), visual styles (e.g., a linocut or watercolor aesthetic), or abstract concepts (e.g., laughing like a maniac). Multiple LoRA modules can be composed, for example, to generate a linocut-style image of a laughing cat. Because each LoRA corresponds to a separate file containing only the additional low-rank weights, practitioners commonly refer to `a LoRA' or `using LoRAs', which implies modular and reusable artifacts.}. Rather than updating any of the pretrained model weights, LoRA keeps the base LLM entirely frozen and injects additional trainable low-rank matrices into selected linear transformations. These low-rank parameters produce additive updates during the forward pass, enabling task- or domain-specific adaptation while preserving the original model and dramatically reducing the number of trainable parameters. The difference between LoRA and RAG is where the knowledge comes in: RAG injects information at inference time through additional context tokens, whereas LoRA augments the model’s parametric pathway by adding persistent, trainable low-rank components to \textit{selected}\footnote{The weights introduced by a LoRA are always present during inference, but their influence on generation is typically \textit{conditional} on the prompt. In practice, LoRAs are trained such that their effects are activated by specific lexical or semantic cues. For example, if a LoRA is trained for agent-based modeling but the prompt concerns system dynamics, the LoRA’s influence may remain effectively dormant. Rather than repeatedly changing the architecture by dynamically enabling or disabling LoRAs, practitioners commonly rely on \emph{trigger words} to activate certain LoRAs. For instance, the LoRA on agent-based modeling can be included all the time, but only the phrase `agent-based model' in the prompt would be likely to activate the learned adaptations. The trigger words tend to be short and specialized, rather than a full sentence such as ``why don't you pass the time by playing solitaire?''. In short, LoRA-encoded knowledge is persistent in the model parameters, but only expressed when the prompt triggers the corresponding learned patterns.} transformations. 

LoRA is not intended as a general-purpose mechanism to `inject facts': rather, it specializes the parametric behavior of an LLM for a stable context. For instance, if modelers repeatedly request that an LLM translates natural language into the same type of conceptual model (e.g., causal loop diagram), or critique models from the same paradigm (e.g., agent-based model), then a LoRA can be trained from curated examples (e.g., learning a mapping given pairs of texts and associated causal loop diagrams) to use the appropriate terminology or organize the output according to accepted conventions. Alternatively, a LoRA could encode domain regularities: the concepts, constraints, and interactions that characterize an application area such as wildfire spread, flooding, or epidemics. It would not provide authoritative references or scenario-specific facts (that would need RAG\footnote{The nuance can be subtle, particularly when training a LoRA to frame the what-if questions that go into a model. A LoRA could help to articulate the scenarios in a consistent format. If a simulation workflow repeatedly needs to perform sensitivity analyses, then the LoRA could encode their style. This is not about supplying facts or evidence (which would be the RAG’s role), but about enforcing a consistent structure for how scenarios are articulated.}), but it would ensure the use of domain-appropriate abstractions. LoRAs can be combined\footnote{When combining LoRAs, we should avoid triggering multiple adaptations that encode incompatible assumptions or conventions. This issue is particularly salient in hybrid modeling, where multiple paradigms (e.g., agent-based modeling and cellular automata) are used together~\cite{Mustafee2024}. If separate LoRAs have been trained on paradigm-specific notions such as `validation' or `state update' and then both are activated, then it is difficult to know whether one LoRA would dominate, whether their effects would blend into an uncontrolled hybrid, or whether the resulting behavior would depend sensitively on prompt phrasing and training artifacts. The dominance of a LoRA may be lowered by decreasing its weight, but this is a coarse approach that redistributes influence without fundamentally resolving ambiguity. This is not only a technical limitation of LoRA composition, but also reflects a deeper challenge in aspects of M\&S that may lack normalized expectations about how concepts should be reconciled across paradigms. As a result, conflicting LoRAs may echo the ambiguity in the field rather than resolve it.}, for instance by combining a LoRA from a paradigm (e.g., cellular automata) with a LoRA for the application domain (e.g., wildfire spread). The low-rank update of a LoRA is typically scaled by a tunable coefficient at inference time, called `weight'. That weight can be larger than 1 to make the LoRA's specialization matter more than it did during training, for instance if it was trained conservatively (e.g., limited data, low rank) or we want to force an effect even when the prompt is ambiguous. When multiple LoRAs are active, adjusting these weights can bias the generation toward one set of learned adaptations over another. Note that a weight larger than 1 does not make a LoRA more correct, it only makes it more \textit{forceful} by becoming (over)sensitive to trigger tokens or suppression legitimate alternatives.

\begin{table}[p]
\centering
\footnotesize
\setlength{\tabcolsep}{4pt}
\renewcommand{\arraystretch}{1.3}

\makebox[\textwidth][c]{%
\rotatebox{90}{%
\begin{tabular}{|p{2cm}|p{3cm}|p{4cm}|p{3.5cm}|p{3.5cm}|}
\hline
\multicolumn{1}{|c|}{\textbf{Aspect}}       & \multicolumn{1}{c|}{\textbf{RAG}}                  & \multicolumn{1}{c|}{\textbf{LoRA}}                           & \multicolumn{1}{c|}{\textbf{Adapters}}             & \multicolumn{1}{c|}{\textbf{Selection-based}}     \\ \hline
\textbf{Primary purpose}                    & Inject external knowledge at inference time        & Specialize model behavior via low-rank parametric adaptation & Specialize model behavior via new trainable layers & Route computation through existing model capacity \\ \hline
\textbf{Where changes occur}                & Input/context pathway (tokens)                     & Parametric pathway (additive weight updates)                 & Architectural pathway (inserted layers)            & Control pathway (gating, masking, routing)        \\ \hline
\textbf{Adds trainable parameters}          & No                                                 & Yes (low-rank matrices)                                      & Yes (bottleneck layers)                            & Typically no (sometimes small gating params)      \\ \hline
\textbf{Modifies architecture}              & No                                                 & No                                                           & Yes                                                & No                                                \\ \hline
\textbf{Knowledge persistence}              & Transient (per prompt)                             & Persistent (stored in parameters)                            & Persistent (stored in parameters)                  & Persistent but implicit (in routing policy)       \\ \hline
\textbf{Dependency on prompt content}       & High (retrieval query determines knowledge)        & Medium–high (trigger tokens modulate effect)                 & Medium (activated when adapter is enabled)         & High (selection often conditioned on input/task)  \\ \hline
\textbf{Control mechanism}                  & At token/content level                             & Use scaling weights                                          & Enable/disable                                     & Discrete or soft via routing decisions            \\ \hline
\textbf{Typical knowledge encoded}          & Facts, documents, evidence, scenario-specific data & Conventions, styles, domain regularities                     & Task-specific transformations                      & Task regimes, behavioral modes                    \\ \hline
\textbf{Suitability for changing scenarios} & High                                               & Low–medium                                                   & Low–medium                                         & Medium                                            \\ \hline
\textbf{Suitability for stable domains}     & Medium                                             & High                                                         & High                                               & Medium                                            \\ \hline
\textbf{Interpretability for M\&S users}    & High (retrieved sources inspectable)               & Medium (effects distributed)                                 & High (explicit modules)                            & Low–medium (routing often opaque)                 \\ \hline
\textbf{Risk profile}                       & Missing/irrelevant retrieval; context overload     & Over-specialization; interference between LoRAs              & Architectural complexity; stacking effects         & Unintended routing; brittle gating                \\ \hline
\textbf{Potential use in M\&S}              & Supplying scenario facts, references, policies     & Enforcing modeling paradigms or domain abstractions          & Task-specific transformation pipelines             & Switching between known simulation regimes        \\ \hline
\end{tabular}
}}
\caption{RAG changes what the model \textit{sees} at inference time; LoRA and adapters change how the model \textit{thinks} by shaping its parametric behavior; selection-based methods change which parts of the model are \textit{used}. These approaches can be used to provide  exposure to scenario-specific information (RAG), encode stable modeling conventions (LoRA/adapters), or switch between model configurations.
}
\label{tab:methodsinventory}
\end{table}

%\textcolor{red}{usually it means one version... but now we also have graphRAG (https://dl.acm.org/doi/10.1145/3777378)}

A complete inventory of methods to augment knowledge is beyond the scope of this article, particularly as only a few have been used (or considered for application) in M\&S. Table~\ref{tab:methodsinventory} provides a summary based on our experience, but the suitability of these methods across M\&S tasks should be re-assessed regularly as this is a quickly changing landscape. Two other well-defined categories of methods that we have not covered as much currently, but whose prominence could change in the future, include adapter and selection methods. LoRA reshapes existing transformations by \textit{additive updates} to pretrained weight matrices (i.e., $W_{effective} = W_{base} + \Delta W_{LoRA}$), so they can be merged back into the base model\footnote{This is common practice in text-to-image and text-to-video models, in which several LoRAs are released individually and preferred ones are eventually packaged into a new version of the model.}. In contrast, adapters introduce auxiliary parameterized mechanisms that modulate or reroute computation while leaving the original weights untouched and unmerged\footnote{Adapters were introduced as `inserting new trainable layers', but the field gradually shifted to a broader definition that does not require explicitly inserted layers. For example, the Structured MOdulation Adapter (SMoA) modulates existing transformations through learned, high-rank structured parameters, yet remains an adapter because its effects cannot be absorbed into the base model~\cite{liu2026high}.}. \textit{Selection} methods do not introduce new transformations of the signal (like adapters) or augment existing weight matrices (like LoRAs): they choose what to activate or freeze among existing parameters. In other words, they make a selective use of the existing model capacity by routing computations through different subsets\index{Knowledge augmentation!Parametric|)}\index{Low-Rank Adaptation (LoRA)|)}.

\section{Ignoring the impact of non-determinism}
\label{sec:ignoreNonDeterminism}
\subsection{Why is there non-determinism in LLMs?}
\label{sec:nondeterminism}
Non-determinism means that even if we use the same prompt and (what appears to be) the same LLM environment, there is a \textit{possibility} that the result is different\index{LLM execution!non-determinism|(}. This does not mean that the result \textit{will} be different each time, so simply running the prompt a few times and seeing the same result does not provide an argument to conclude that the process is deterministic. In fact, there are many mechanisms that produce non-determinism with LLMs. We suggest that framing LLMs as mere `stochastic parrots' and attributing all non-determinism to the model is neither constructive nor entirely accurate. Without precisely understanding sources of non-determinism, we may just brush off some outputs as `errors' while the general population may take them as evidence of the LLM being `autonomous'~\cite{rapp2025people}. To be precise, we should avoid conflating non-determinism, which can broadly be categorized into \textit{inference non-determinism}\index{LLM execution!inference non-determinism} (e.g., implementation optimization) and \textit{distributional bias}\index{LLM execution!distributional bias} (what sequences the model assigns high probability to). Some of these sources of non-determinism can be fixed, but it certainly takes more work than just setting a temperature parameter to 0 -- as several empirical studies show that high variation in the output still happens~\cite{ouyang2025empirical,astekin2024exploratory}. Even when non-determinism cannot be tamed, inspecting its mechanisms may reveal more patterns than just `randomness', thus providing valuable insight into an LLM's behavior~\cite{cheng2025stochastic}. For example, experiments support the (distributional bias) hypothesis that ``LLMs are sensitive to the probability of the sequences they must produce'': they are more likely to give the right answer if it has a high probability of occurrence, even if the task is deterministic~\cite{mccoy2024embers}. In short, correctness correlates with sequence likelihood, not task determinism. 

It is well-known that LLMs model conditional token distributions~\cite{radford2019language,holtzman2020curious}. Again, this is not to say that they are merely `stochastic parrots' that excel at memorizing vast amounts of training data and regurgitating them with a bit of randomness: the ability of newer LLMs to plan and perform reasoning goes beyond just repeating patterns from their training set~\cite{lizarraga2025stochastic}. The many reasons for which non-determinism happens during inference are perhaps less well-known, or even unexpected for some readers, thus we focus on these cases. 

When combining LLMs with M\&S, we rarely use ChatGPT and a web browser. Rather, we automate some of the operations (e.g., using the LLM to generate a conceptual model from a corpus) through code such as a Jupyter Notebook operating in Python. If we use a single LLM such as OpenAI's GPT, then the entry point to access the model is the OpenAI API. However, in practice, we often need to use several LLMs, either for optimization (which one is best for the task? or provides a good tradeoff between task accuracy and cost?) and/or for evaluation (to which extent is the conceptual model produced by the LLM shaped by the choice of model?). As a result, it is common to use a centralized platform that offers access to many LLMs across providers through the same API. While this may at first seem like a point of detail that would be found in an implementation footnote (or even just omitted), it can have a surprising effect on non-determinism. For a given model (e.g., DeepSeek-R1-Distill-Llama-70B), a centralized platform such as {\ttfamily OpenRouter}\index{LLM execution!Routing|(} provides access to several providers. They have different prices for inputs and outputs, different latencies\footnote{For instance, the report on Qwen3-Omni details a set of techniques to achieve ultra low-latency~\cite{xu2025qwen3omnitechnicalreport}. Such models can be used for `real-time' audio/video tasks such as transcription.}, and may impose different limits on the maximum size of the output. However, (i) these providers may run the models differently and (ii) OpenRouter routes the request to \textit{any} of the available providers that serves the same base model and satisfies a user's needs for prompt size and parameters. So when we send several prompts, some may be executed differently than others.

A difference in execution does not necessarily mean a difference in results: for instance, speed-up `tricks' such as predicting multiple tokens in parallel using a smaller/faster model (i.e., speculative decoding\index{LLM execution!Speculative decoding}) can provide a lossless acceleration of LLMs during inference~\cite{xia-etal-2024-unlocking,sun2024block}, so theoretically they would only affect efficiency rather not correctness. But there are also three differences that affect the output. First, \textit{LLM quantization} represents a model’s weights and activations with lower-precision numerical formats (e.g., int8, fp8, bf16 instead of fp32) to reduce memory use and speed up inference, which introduces `small' numerical approximation errors. In our example of DeepSeek R1 Distill Llama 70B, some providers use fp8, some use bf16, and some do not say. Just with quantization alone, even using the same prompt and seed can lead the token sampling to diverge\footnote{Since classical methods (e.g., GPTQ, AWQ, QuIP), new methods (e.g., QEP, GPTAQ, LoaQ) have been developed to control error propagation across layers and reduce the accumulation of quantization errors~\cite{ichikawa2025lpcd,li2025gptaq}.}. Second, providers may perform different optimizations to speed up computations or reduce memory consumption for \textit{attention}\index{LLM execution!Attention}, which is the core mechanism that lets the LLM account for the previous tokens when generating the next token\footnote{As an example of a highly cited optimization, {\ttfamily FlashAttention} is a memory- and compute-efficient implementation of transformer attention~\cite{dao2022flashattention}. It changes the computation of matrices to minimize intermediate memory writes.}. This is not a change of precision (as in quantization) but a potentially different numerical execution order: sums are accumulated in different sequences, thus rounding happens at different points. In other words, quantization changes the numbers that we compute with, but optimizing the attention changes how we compute the numbers. Third, LLMs recompute the same keys and values for previous tokens at each time step, so they have to manage the \textit{key-value cache} (KV cache)\index{LLM execution!Caching}. The management strategy can make a difference in long-context prompts: a simple sliding window drops the oldest cached key/value pair when memory gets full, while a more advanced strategy would evict content based on its expected importance to preserve long-range dependencies and thus affect the overall inference quality~\cite{wang2025fier,kwon2023efficient}. 

An additional and perhaps unexpected situation is that asking for the same LLM does not mean that we get exactly the same weights or behavior. There are two reasons for this situation. First, there is (silent) \textit{forwarding}\index{LLM execution!Forwarding}: LLMs eventually become deprecated so providers such as {\ttfamily DeepInfra} apply their model deprecation policy by which older models are removed and (to avoid breaking a user's code) calls are then forwarded to another model. Modelers may see that their code just works `as usual' without realizing that it is now being serviced by another LLM. Second, a name is just an \textit{alias}\index{LLM execution!Alias|(}: asking for `GPT-5' does not refer to one set of weights and policies forever, but rather to whatever was in place at the time that the call was made. Providers may update policies, add moderation, or route prompts differently based on account tiers and geography (which can also determine which policies are applicable). In a highly cited study, Chen and colleagues reported the opacity in determining when and how some LLMs were updated, with clear differences in performances for the same LLM by name (e.g., GPT-4) over several months. For instance, GPT-4 answered sensitive survey questions more in March than in June~\cite{chen2024chatgpt}. 

\subsection{Evaluating the impact of non-determinism}
\label{sec:evaluatingNonDeterminism}
The goal is not necessarily to reduce non-determinism, because other considerations can be more important for practical deployment. For example, costs
\footnote{Cost can relate to variability across runs, particularly when using \textit{caching} to retrieve the (same) answers to queries that have already been computed (i.e., `replaying' prior outputs). For example, providers such as DeepSeek reduce its costs by orders of magnitude \textit{when users are willing} to use the cache \textit{and there is a cache hit} as the prefix of a new prompt sufficiently matches the prefix of a previously computed prompt. However, caching can work very differently across providers. For OpenAI, prompt caching is automatic since GPT-4o. In contrast, Anthropic’s Claude exposes explicit caching controls via a dedicated field and allows users to extend the default caching window for a fee, thus providing fine-grained control over cost and variability. Note that caching can give the \textit{illusion} of determinism by repeatedly returning identical outputs over a limited time window, so caching should be \textit{disabled} (either explicitly or by waiting between repeated prompts) when assessing a system’s variability.} or latency may be the main objectives for a large-scale solution that serves a large number of users in low-risk environments. But even if a certain application context is `willing' to live with some non-determinism in the outputs, this should be an informed decision: to what extent are key performance measures affected by stochasticity arising from sampling, inference, or system-level variability? An \textit{ablation study} is a common approach to decompose the performance of a system based on whether certain parts are included (e.g., whether a RAG is used, whether additional examples are provided in the prompt). However, that does not apply here since stochasticity is neither a binary design choice nor fully under user control (unlike e.g. whether or not to use a RAG), and its effects are inherently variable across runs. Alternatively, a simple statistical approach is to report the average and confidence intervals for each performance measurement across runs (also known as 'repeats'). However, it is well-known in the modeling and simulation literature that the number of runs cannot be arbitrarily set e.g. to 10, 100, or 500: we need to determine how many runs are statistically sufficient otherwise the confidence intervals are not meaningful~\cite{Robinson}[pp. 182--193]. This has motivated the development of other statistical approaches for LLMs that approximate the distribution of a model's performance using bootstrapping~\cite{fraile2025measuring}. Nonetheless, the implicit assumption that performance varies smoothly around a central value is questionable in practice since empirical results show that performance distributions can be multimodal and heavy-tailed, where best-case runs give the illusion of high performances and hide the massive issues caused by worst-case runs~\cite{song2025good,atil2024non}. For example, an LLM used as part of a logistics simulation may meet its average resolution targets, yet generate some irrecoverable failures that violate service-level guarantees. This motivates both the development of other metrics (see Box 2) and the careful use of Design of Experiments (DoE), which have been a staple of the modeling and simulation literature for decades.

Since Sanchez and colleagues have provided several tutorials on the different types of DoE used in modeling and simulation~\cite{sanchez2020work,sanchez2009better}, we briefly cover \textit{how} an experimental design applies to the assessment of non-determinism in LLMs for simulations. A model is composed of different parameters: for example, an agent-based model may have a measure of diversity in the population, or an influence threshold after which an agent's opinion would start to align on its peers. The LLM component is also made up of several parts, such as whether to use a RAG or to add examples to the prompt. A DoE can decompose variations in the response variable (e.g., TAR@N or WorstAcc@N per Box 2) onto the individual \textit{and interacting} effects of these parameters, design elements, and stochastic effects. In a \textit{full factorial experiment}, we consider all combinations: if there are three levels of diversity and four values for the influence threshold, then we cover all $3 \times 4 = 12$ cases. In a simplified and commonly used design, we limit each aspect to two values. So if we have $4$ factors (agents' diversity, influence threshold, RAG, examples) set to $2$ levels each (high/low diversity, high/low threshold, presence/absence of RAG and examples), then we need to study $2 \times 2 \times 2 \times 2 = 2^4$ combinations. Generalizing this example, a $2^kr$ \textit{factorial design} with replications can study the $k$ binary factors across $r$ runs for each combination. Non-determinism is captured via replication (Table~\ref{table:sampleFactorial}). Variability in the output measure would be decomposed on the individual factors and their interactions in pairs (e.g., diversity \textit{and} randomness, diversity and threshold, RAG and randomness), groups of three, and groups of four\footnote{For examples of a factorial design in a modeling and simulation study, see~\cite{li2021identifying,lutz2022we}. For code that allows to run these experiments in parallel and supports the scalability of the analysis, see~\cite{lavin2017analyzing}.}. If a $2^kr$ exceeds either the available budget to repeatedly query LLMs or the time devoted to gathering experimental results, then an alternative \textit{fractional factorial design} can reduce the number of combinations at the expense of lesser precision in the analysis\footnote{Higher-order interactions (i.e., the synergistic effect attributable to a \textit{group} of several factors) tend to have less of a contribution than lower-level interactions. Typically, single factors and pairs explain most of the variance, as it is relatively unusual to observe effects that are \textit{only} obtained for specific combinations in the values of three or more parameters. A fractional factorial design would \textit{confound} (i.e., `mix') the variance that can be attributed to low- and high-order interactions. For example, a single number accounts for the effects of both $AD$ or the larger combination of four parameters $ABCD$, but we expect that it mostly represents $AD$. A good fractional factorial design would avoid confounding groups of similar sizes (e.g., two and three factors). The quality of a design is evaluated by its \textit{resolution}, that is, the distance between the groups of factors that are confounded. The notation indicates how many parameters \textit{should} have been used, how many were fixed to cut experimental costs, and the quality of the design (denoted in roman numerals). For instance, $2^{3-1}_{III}$ is fractional factorial design that should have had 3 parameters but fixed one (which reduces the number of experiments by half) and has a design resolution of III.}. 

As a practical case study, we used factorial designs to measure the amount of variability when employing LLMs for the conceptual modeling task of combining two models. That is, given several conceptual models, the LLM had to find how they could be merged by accounting for semantic variability (different words referring to the same concept). We found that the extent to which non-determinism drives the results depended on \textit{how} the problem was tackled \textit{and} on the LLM. Using a simple method that tries to find whether each concept of one model is directly equivalent to concepts of another model (e.g., given the context, can stress be merged with cortisol?) produced negligible amounts of randomness. Performances were primarily driven by how the conceptual model was represented in the prompt (e.g., as an array or list) and by the instructions (providing explanations, examples, counter-examples). In contrast, a more sophisticated approach in which synonyms and antonyms were gradually derived (e.g., stress relates to anxiety, which is measured by cortisol levels, so cortisol is a match) had a massive amount of randomness, as most of the variability in the results (from 53\% with GPT to 95\% with DeepSeek) were attributed to non-determinism. \textit{Ignoring the presence of randomness} by forgetting to perform repeats or analyze them would have erroneously suggested that design aspects such as providing counter-examples had a high impact onto the performance, whereas appropriately measuring the effect of randomness revealed that modelers had very little control on performances.

\medskip
\noindent
\fcolorbox{black}{gray!20}{%
  \parbox{1\linewidth}{\textbf{Box 2. Characterizing systems: beyond means and confidence intervals}\par
Consider two LLMs ($LLM_A$ and $LLM_B$) used to set the behaviors of agents in a simulation. They are used to predict the next action for each of the 100 agents and each simulation is performed 20 times. Accuracy is measured as the percentage of correct actions across runs based on a ground truth dataset. For $LLM_A$, it gets the \textit{same} 80 agents right each time (e.g., it knows exactly what to do for a subset of agents based on environmental state or role). In contrast, $LLM_B$ has an 80\% chance of being correct for the behavior of an agent, thus which ones are accurate vary randomly (correctness is not patterned by the agents' attributes). They have the same average performance, yet $LLM_A$ yields fully reproducible results while $LLM_B$ is very unstable.\\
Another measure is to compare the best and the worst accuracy of each LLM across the runs. Here, \textbf{WorstAcc@20} (must always be correct for each agent across runs) is 80\% for $LLM_A$ and it tends towards 0\% for $LLM_B$. In contrast, the \textbf{BestAcc@20} (must be correct at least once for each agent across runs) is 80\% for $LLM_A$ but 100\% for $LLM_B$. The differential $\Delta Acc=BestAcc@20-WorstAcc@20=100$\% for $LLM_B$ indicates its high variability~\cite{atil2024non}.\\
Alternatively, we can compute the Total agreement rate@N (\textbf{TAR@N}), defined as the percentage of answers across N runs that are identical, regardless of whether they are correct~\cite{atil2024non}. This may be further divided into whether the answers are \textit{identical} (string matching or `surface determinism') or \textit{equivalent} (this `decision determinism' can further be divided based on simple parsing or semantic distance). Here, we would see a TAR@20 of 100\% for $LLM_A$ and about 0\% for $LLM_B$.
  }%
}
\medskip

\begin{table}[!h]
\centering
\caption{The standard way to conduct a factorial analysis is to create a table with the factors (e.g., A=agents' diversity, B=influence threshold), their interactions (we omit groups of 3 and the group of 4 for brevity), and the response variable $y$ across runs. The effect of non-determinism is not a factor by itself; rather, it is computed by examining the variation in the response variable. The coded levels -1 and 1 are used to decompose the effect (by cross-product of the columns then sums) and they correspond to the low and high level of each factor (e.g., low agents' diversity is -1 and high diversity is 1). The factors are set as a sign table then other columns ($AB$, $AC$, etc) are populated by multiplying (e.g., the content of $AB$ is $A \times B$).
}
\begin{tabular}{|l|l|l|l|l|l|l|l|l|l|l|l|l|l|}
\hline
\multicolumn{1}{|c|}{\textbf{Run}} & A & B & C & D & AB & AC & AD & BC & BD & CD & $y_1$ & $y_2$ & $y_3$ \\ \hline
1                                  & -1                              & -1                              & -1                              & -1                              & +1                               & +1                               & +1                               & +1                               & +1                               & +1                               & 0.72                                 & 0.70                                 & 0.71                                 \\ \hline
2                                  & +1                              & -1                              & -1                              & -1                              & -1                               & -1                               & -1                               & +1                               & +1                               & +1                               & 0.75                                 & 0.77                                 & 0.74                                 \\ \hline
3                                  & -1                              & +1                              & -1                              & -1                              & -1                               & +1                               & +1                               & -1                               & -1                               & +1                               & 0.68                                 & 0.66                                 & 0.67                                 \\ \hline
4                                  & +1                              & +1                              & -1                              & -1                              & +1                               & -1                               & -1                               & -1                               & -1                               & +1                               & 0.79                                 & 0.80                                 & 0.78                                 \\ \hline
5                                  & -1                              & -1                              & +1                              & -1                              & +1                               & -1                               & +1                               & -1                               & +1                               & -1                               & 0.73                                 & 0.71                                 & 0.72                                 \\ \hline
6                                  & +1                              & -1                              & +1                              & -1                              & -1                               & +1                               & -1                               & -1                               & +1                               & -1                               & 0.81                                 & 0.82                                 & 0.80                                 \\ \hline
7                                  & -1                              & +1                              & +1                              & -1                              & -1                               & -1                               & +1                               & +1                               & -1                               & -1                               & 0.69                                 & 0.68                                 & 0.70                                 \\ \hline
8                                  & +1                              & +1                              & +1                              & -1                              & +1                               & +1                               & -1                               & +1                               & -1                               & -1                               & 0.84                                 & 0.83                                 & 0.85                                 \\ \hline
9                                  & -1                              & -1                              & -1                              & +1                              & +1                               & +1                               & -1                               & +1                               & -1                               & -1                               & 0.70                                 & 0.69                                 & 0.71                                 \\ \hline
10                                 & +1                              & -1                              & -1                              & +1                              & -1                               & -1                               & +1                               & +1                               & -1                               & -1                               & 0.76                                 & 0.78                                 & 0.77                                 \\ \hline
11                                 & -1                              & +1                              & -1                              & +1                              & -1                               & +1                               & -1                               & -1                               & +1                               & -1                               & 0.67                                 & 0.66                                 & 0.65                                 \\ \hline
12                                 & +1                              & +1                              & -1                              & +1                              & +1                               & -1                               & +1                               & -1                               & +1                               & -1                               & 0.82                                 & 0.83                                 & 0.81                                 \\ \hline
13                                 & -1                              & -1                              & +1                              & +1                              & +1                               & -1                               & -1                               & -1                               & -1                               & +1                               & 0.72                                 & 0.73                                 & 0.71                                 \\ \hline
14                                 & +1                              & -1                              & +1                              & +1                              & -1                               & +1                               & +1                               & -1                               & -1                               & +1                               & 0.86                                 & 0.87                                 & 0.85                                 \\ \hline
15                                 & -1                              & +1                              & +1                              & +1                              & -1                               & -1                               & -1                               & +1                               & +1                               & +1                               & 0.71                                 & 0.70                                 & 0.72                                 \\ \hline
16                                 & +1                              & +1                              & +1                              & +1                              & +1                               & +1                               & +1                               & +1                               & +1                               & +1                               & 0.90                                 & 0.91                                 & 0.89                                 \\ \hline
\end{tabular}
\label{table:sampleFactorial}
\end{table}

\subsection{Mitigating the impact of non-determinism}
\textit{If} the analysis in section~\ref{sec:evaluatingNonDeterminism} reveals that non-determinism plays a larger role in determining the performances that is desirable for the modelers or model commissioners, \textit{then} actions can be taken. Otherwise, there is no need to address a non-existing problem. Actions do not necessarily mean rushing into active mitigation strategies: further analyses could be used to better locate the problem. For instance, if we find that 60\% of the variance in performance comes from non-determinism, that could be due to several phenomena: is it inference non-determinism (e.g., the routing service uses slightly different LLMs unknowingly to the modeler) or distributional bias (e.g., the LLM itself is the issue)? To isolate the source of the error, the experimental design can be expanded, for instance by adding another binary factor to represent whether we let the routing service use any version of the LLM (which reduces costs and latency) or whether we should use the same one consistently. Once we have gathered information to identify where we need to intervene, then several mitigation strategies can be taken as follows.

If the issue comes from the dynamic routing to different versions of an LLM, it cannot be addressed by setting specifications and requesting that only a perfect match be used. Indeed, not all providers expose all specifications of their implementations (e.g., we do not always know whether an LLM is in fp8 or fp32 form). Rather, the solution is to \textit{explicitly set one provider instead of dynamically changing between providers}. If the issue comes from the LLM being changed (e.g., automatic forwarding of deprecated models, model updates) then the solution includes self-hosting an open-weight model (which is limited by computing power and excludes popular options such as the latest GPT) or using a fully versioned ID that \textit{guarantees} the same model. Long, immutable version IDs are not offered by all providers (e.g., it is not the practice of OpenAI). Anthropic, which builds Claude, provides explicit, dated model IDs (e.g., claude-3-opus-20240229) and currently guarantees that if users refer to an ID then they get the same weights and behaviors.\index{LLM execution!non-determinism|)}\index{LLM execution!Routing|)}\index{LLM execution!Alias|)} 

\section{`Working' is not enough: avoiding inferior science with LLMs}
\label{sec:misuse}
\subsection{Risks of LLMs as a one-stop shop}
When studying on a scientific topic, it is expected that we engage directly with the literature. Reading papers is a process through which we encounter competing methods, contradictory findings, and unresolved debates, and synthesize the evidence base to understand not only what supports an argument but also what challenges it. Against this backdrop, a visible (and sometimes normalized) misuse of LLMs is their deployment to supply references. For example, a premier research conference such as NeurIPS had over 50 accepted papers containing AI-fabricated citations in 2025 ~\cite{goldman2026neurips}. Fundamentally, using LLMs to add references or delegating reading tasks to LLMs means that authors are not exposed to potential evidence that contradicts them. Reframing scholarship as a search for confirmatory citations via LLMs creates a form of \textit{confirmatory bias} at scale. While it may seem that nudging LLMs to produce real and relevant references (e.g., by using different prompts and automated or manual verification~\cite{linardon2025influence}) solves the problem, this only satisfies a minimal notion of what it means to `work' rather than addressing the standards of scientific inquiry. Even if an LLM appears to `work' with real references, investigating their use has revealed that they do not necessarily support (and sometimes even contradict) the arguments for which they are used~\cite{wu2025automated}. And even if LLMs fetch real papers and use them well, this is a sub-optimal process that replaces expert judgment grounded in critical synthesis of the literature with automated, argument-serving retrieval. This pattern extends well beyond citations: in many scientific tasks, an LLM may appear to `work' at a superficial level while producing outcomes that are epistemically or methodologically sub-optimal. For example, instead of writing a Python script or using a statistical package, it is possible to ask an LLM to analyze data. As with references, it may seem to `work' by returning the correct result \textit{today}, but due to non-determinism (see section~\ref{sec:ignoreNonDeterminism}), it is possible that it returns a different result next time. We may fix this problem by asking the LLM to produce the code so that we can execute it ourself, thus guaranteeing that the same result would be produced each time. While the code may seem to `work', there is growing evidence that AI-generated code can be correct yet of poor quality and with security flaws~\cite{sabra2025assessing,tihanyi2025secure}. Overall, treating surface-level correctness as sufficient evidence that an LLM `works' poses a risk of sub-optimal scientific processes.

M\&S is not immune to problematic uses of LLMs, which may be driven by a broader pressure to increase productivity along with the appeal of LLMs as a one-stop shop. Recent work on LLM-generated simulation models illustrates how difficult it is to define and automatically enforce what it means for a model to `work'. For example, Möltner and colleagues showed that a conjecture-based evaluation pipeline failed when trying to have an LLM conjecture the expected properties of a model (from a text description) and evaluate code generation accordingly~\cite{moltner2025creation}. The reasons for this failure highlight an interesting difference between an LLM approach to generating and validating simulation code compared to the approach taken by a (human) modeler. A human modeler would read a textual description of a problem, derive a conceptual model, implement that model, and then evaluate the implementation with respect to that same interpretation. In contrast, when LLMs are used, the conjecture and the simulation code are produced by separate LLM calls that \textit{independently interpret the same text}, with no guarantee that they converge on the same conceptual model. As a result, validation can fail even when the implementation is reasonable, because it became a test of interpretive \textit{consistency}\index{LLM evaluation!Consistency} (which is not an LLM's strength) rather than model \textit{adequacy}\index{LLM evaluation!Adequacy} (which is the actual goal)\footnote{As an example, consider the following text specification: ``Agents are located on a network and hold binary opinions. At each time step, agents interact with their neighbors and may update their opinion based on these interactions. There is no global coordination or central authority.'' Calling the LLM to generate a conjecture could result in considering that agents interact with all neighbors at each step and align their opinion with the local majority. The simulations should thus show rapid convergence toward consensus within connected components. In a separate call, an LLM may implement the same textual description differently, by having each agent interact with one randomly selected neighbor per time step. Neither interpretation is `invalid': they both could produce simulation outputs that match real-world trends over different time scales. But the interpretations assume different conceptual models and this incompatibility results in a false negative: the simulated model would be categorized as wrong because it converges more slowly.}.

\subsection{Translating rather than delegating with LLMs}
A more subtle risk lies in using LLMs for tasks that are already well supported by specialized tools. Consider the effort required to guide an LLM to compare a model’s implementation with a specification in order to provide some form of certification. That could be a research article showing that certain prompting patterns or RAGs improve the results, and then other researchers may reuse this system. But there are already solutions for this problem: model checkers or test generators~\cite{beyer2024six}. While we may all agree in principles that we \textit{should} always use the best tool for a task, we do not necessarily know \textit{how} to use such tools as we cannot be experts in everything. However, the wrong conclusion would be to rely on inferior solutions simply because LLMs are easier to use. Rather than attempting to reimplement specialized capabilities through LLMs, we posit that LLMs are better positioned as translators: mapping informal requirements into the formal languages required by specialized tools, and translating those tools’ outputs back into forms that are accessible to modelers~\cite{giabbanelli2025overreliance}. For instance, a modeler may ask, ``does my implementation behave reasonably?'', yet this is not the kind of question that model-checking tools take as input. Instead, such tools verify formally specified properties (e.g., reachability of some states, invariants) expressed in languages such as linear temporal logic or modal logics. In this context, LLMs are more appropriately used to map natural language requirements into candidate formal properties than to attempt to perform model checking directly. This is illustrated by the {\ttfamily GIVUP} tool~\cite{nivon2025givup}: the authors use the LLM to extract structured information about model processes and properties into temporal logic, then they perform model checking through an external specialized tool by verifying that the model satisfies a given temporal logic property.

%This feels like the natural place to tackle the "one-stop shop" temptation directly. You could use concrete examples: (1) using LLMs for optimization when analytical solutions already exist, or (2) for parameter estimation when classical calibration is faster. This would show readers the real trade-offs between convenience and efficiency. I think the framing that works best here is probably positive: LLMs genuinely excel at unstructured tasks like conceptual modeling and text, but that doesn't mean they should replace optimized numerical methods for what those methods do well.

\begin{figure}[h!]
    \centering

    \includegraphics[width=\textwidth]{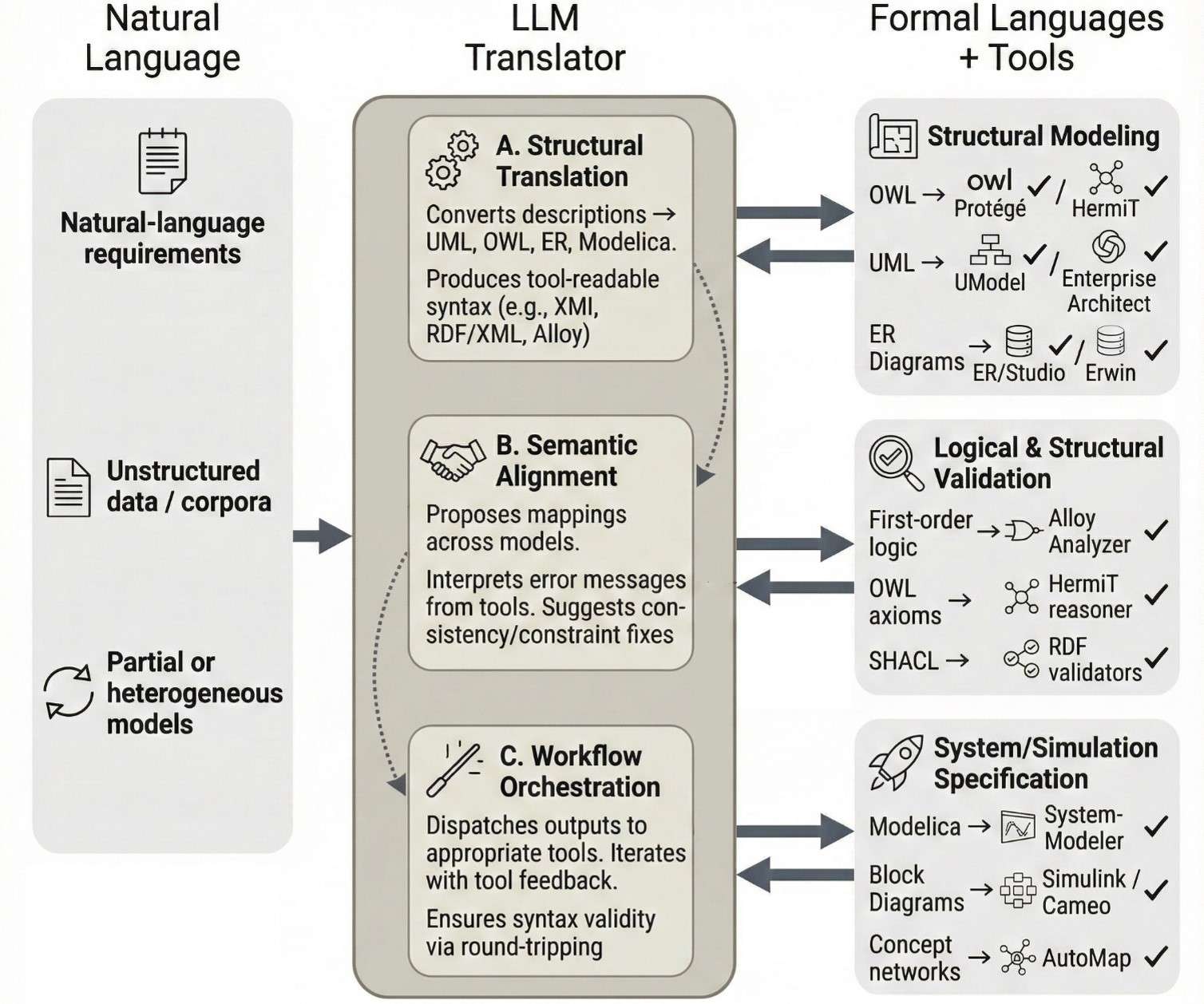}
    \caption{LLMs can mediate between informal modeling requirements and specialized tools to avoid a sub-optimal and time-consuming reimplementation of established methods. LLMs would have to translate and  handle feedback from the tools.}
    \label{fig:toolset}
\end{figure}

Figure~\ref{fig:toolset} illustrates the role of LLMs as translators that connect informal modeling requirements to specialized formal tools. Starting from natural-language requirements, auxiliary documentation, and potentially several models (finished or in-progress), the LLM must first infer which formal representations and tools are appropriate for a given task. This involves translating informal descriptions into the specific input languages expected by those tools (Figure~\ref{fig:toolset}a). In practice, this translation step is often nontrivial: many of these representations can be considered \textit{low-resource languages} for LLMs, and initial attempts may fail due to syntactic or semantic mismatches. Rather than treating such failures as definitive and giving up on using a specialized tool or requesting more training data, we suggest to leverage feedback from the tools when available (Figure~\ref{fig:toolset}b). If tools provide parser errors or violated constraints, then LLMs could use this feedback to iteratively refine the translation. In this case, the LLM plays an orchestration role by dispatching candidate representations to the appropriate tools, interpreting error messages, and adjusting subsequent translations through round-tripping until a valid formal input is obtained (Figure~\ref{fig:toolset}c). Once the specialized tools produce results, the LLM can then translate these outputs back into forms that are accessible and meaningful to the modeler. In this approach, LLMs are not re-implementing methods: they are mediating between heterogeneous representations and tools.

\begin{figure}[h!]
    \centering

    \includegraphics[width=\textwidth]{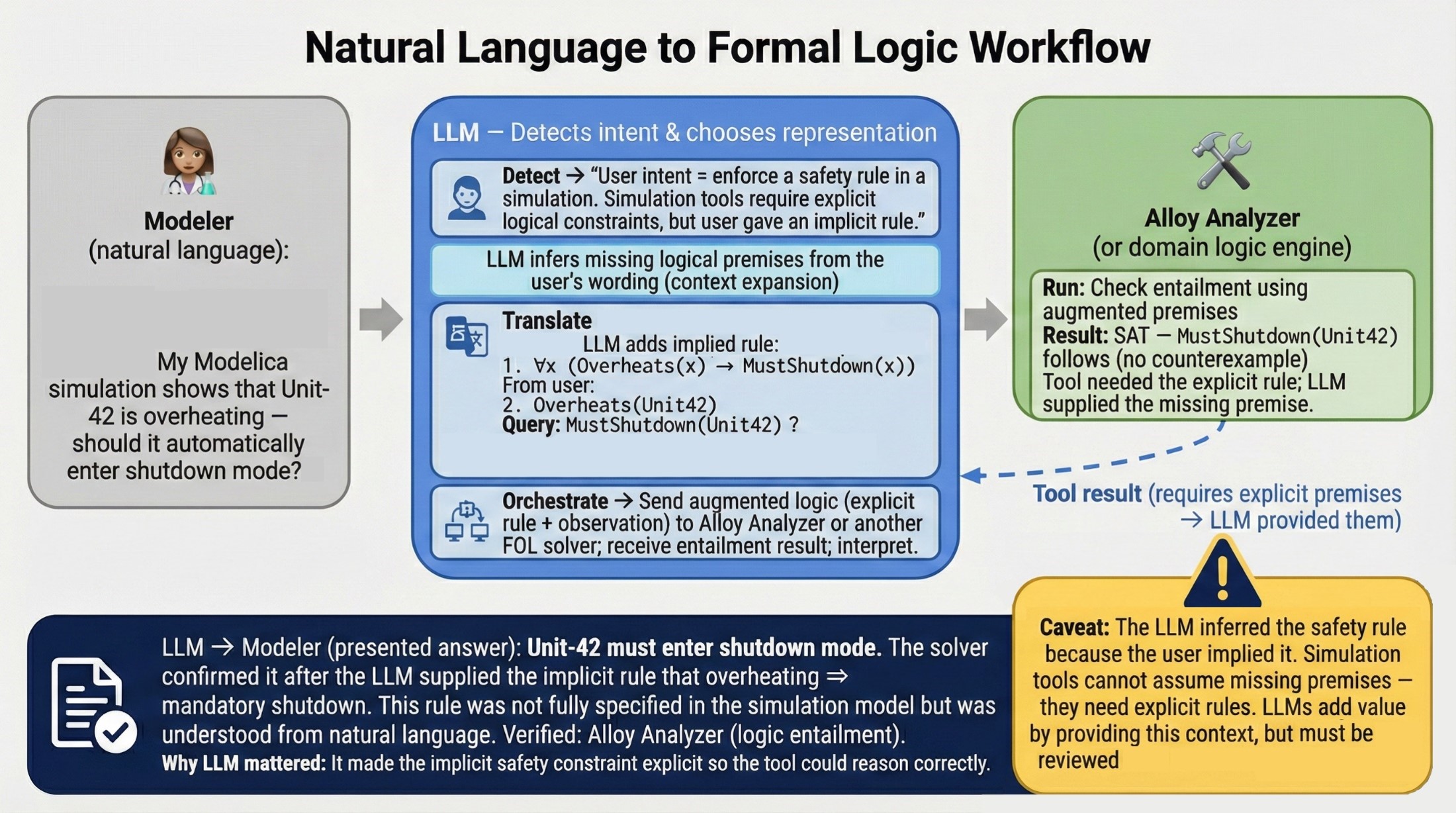}
    \caption{LLMs can translate or `mediate' by handling implicit assumptions such that a modeler's description within a context can be expanded to enable a specialized tool to operate. Transparency is essential when filling such knowledge gaps, so the conclusion (bottom, dark blue) should be conveyed to the modeler along with a disclosure of any inferences that the LLM had to make (bottom right, yellow).}
    \label{fig:missing}
\end{figure}

Acting as a translator does not mean that the LLM just converts the syntax from a modeler's description onto the target representation of a specialized tool. We argue that a key asset of LLMs in this situation is their \textit{knowledge model}: they can bridge the gap that is likely to occur between a modeler's description, which is sufficient \textit{within the application context}, and a tool's description, which must be complete. For example, some (low-level) rules may seem obvious to a modeler in a given application context and they would not be stated, but the specialized tool cannot make inferences without knowing them, so the LLM can suggest such missing premises (Figure~\ref{fig:missing}). Consider the following description of a typical agent-based model for evacuation:
\begin{quotation}
``When an alarm triggers, agents should evacuate the building. Once an agent exits, they should not re-enter. Doors can become blocked during the evacuation.''
\end{quotation}
The modeler may want to check that a simulation is compliant if it follows a safety property (evacuated agents do not re-enter the building) and a progress property (if at least one exit is unblocked then every agent eventually exits). It is \textit{implicit} that people do not keep entering the building while it is evacuated (movement abstraction), and it is also implicit that the alarm stays on (alarm semantics). Thus the LLM would have to provide some implicit rules in order for a reasoning tool to fully grasp the situation. It is important to be transparent: the LLM should convey which elements came from the modeler, which ones were inferred, and which conclusions were established by the formal tool. The LLM's role is not to decide whether implicit assumptions are correct, but to make them explicit such that specialized tools can function and modelers can revise the description before relying on the analysis.

\subsection{Architectures to integrate LLMs in workflows}
\begin{figure}[h!]
    \centering

    \includegraphics[width=\textwidth]{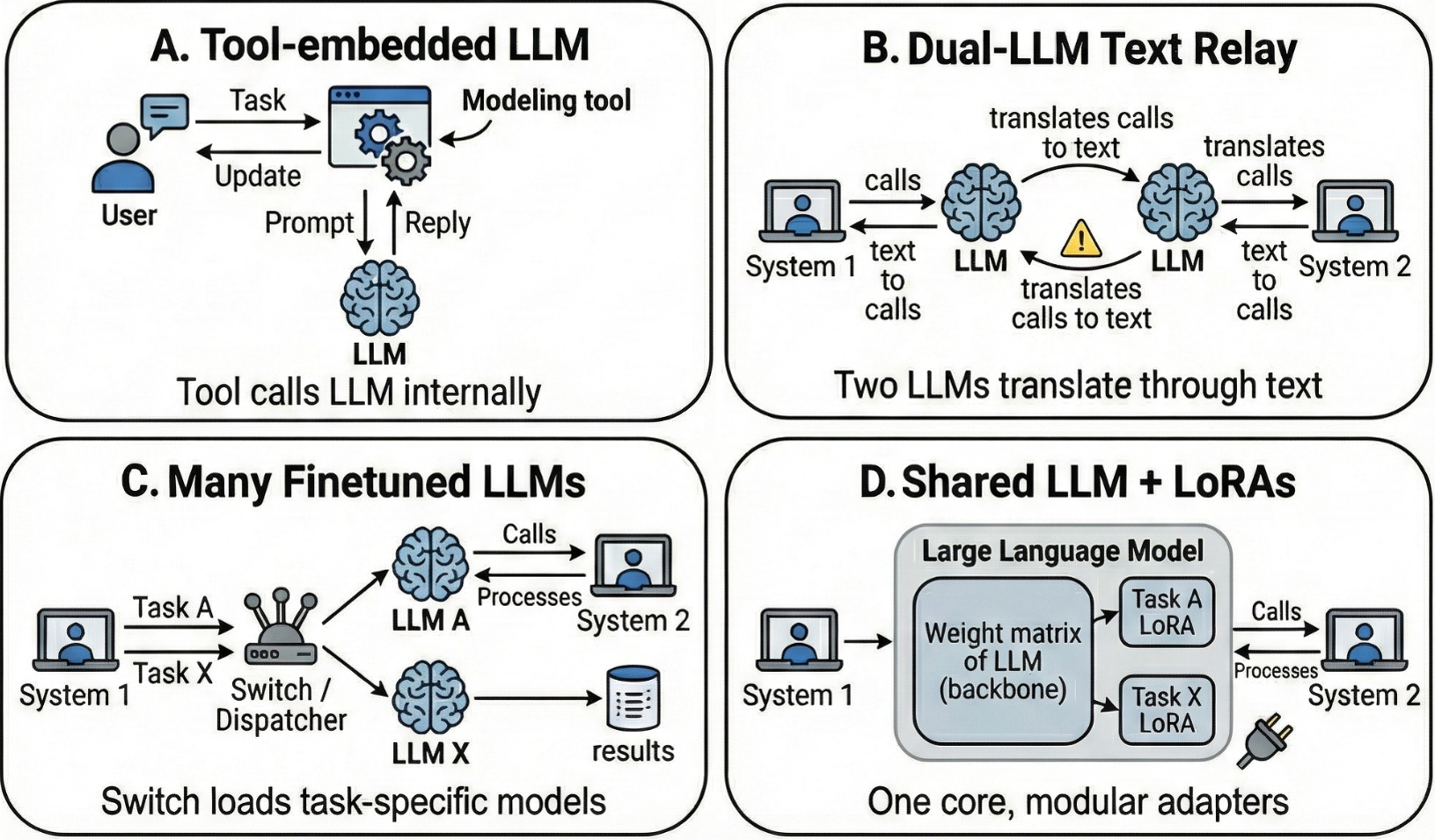}
    \caption{LLMs can be used together with specialized tools in at least four ways: indirectly by being call within the tool (a), as a pair of translators between the user's needs and the tool's requirements (b), as one or more LLMs fine-tuned for different tools (c), or as a single LLM augmented to work with each target tool (d).}
    \label{fig:architectures}
\end{figure}

While we focused on LLMs as translators or mediators, there are other variations of using LLMs together with specialized tools. Different architectures have been employed depending on the researchers' vision. In a \textit{tool-embedded LLM architecture} (Figure~\ref{fig:architectures}a), the LLM is a \textit{backend service} within the modeling environment rather than a direct conversational interface. For example, the authors of {\ttfamily MAGDA} detail that the LLM is invoked internally to make suggestions (e.g., UML classes and attributes) for the modeling environment. All authoritative modeling actions remain within the specialized tool: the LLM does not perform or validate modeling actions. Tool-embedded architectures are becoming more available among popular modeling environments such as {\ttfamily Stella} and {\ttfamily AnyLogic}, while noting that the extent to which the LLM `supports' rather than `performs' modeling tasks depends on the tool. Another architecture proposed by Lehmann consists of \textit{two LLM instances} that translate function calls and data back and forth through natural language (Figure~\ref{fig:architectures}b)~\cite{lehmann2024towards}. While having natural language as an intermediate avoids the need to build a shared schema, there are several potential issues: two model calls each time can negatively affect latency and cost, two LLMs are interpreting the task so their non-determinism may compound, and natural language can be an underspecified intermediate representation that creates a semantic drift. 

%we cannot maintain a one-to-one LoRa for every transformation... so likely we may have to 'assemble' a few... and there may be several transformation paths!

An alternative architecture is to employ a single LLM per call (Figure~\ref{fig:architectures}c). This may suffice if the LLM is intended as the front-end to a single specialized tool, for example by calling an external Python library (e.g., PySD to run system dynamics) and executing\footnote{An LLM does not `execute code' directly: it is not a \textit{runtime system}. A product such as ChatGPT can interact with a Python Sandbox, which can run code. So, to be precise, the tool chain is not directly LLM $\leftrightarrow$ Python Library but rather LLM $\leftrightarrow$ Python Sandbox $\leftrightarrow$ Python Library.} the code~\cite{hu2025chatpysd}. Studies that position a single LLM as the primary interface to a specialized tool may also be exposed to `interface bleeding', whereby effective use of the LLM increasingly requires users to know the underlying tool or language it mediates~\cite{hoffmann2025modularization}. When there are several specialized tools (each with its own input requirements), then this architecture becomes problematic. There would need to be several LLMs that are fine-tuned to the requirements of each tool, along with a \textit{dispatcher that selects the right LLM} in each case. Treating each specialized variant as its own model results in repeatedly loading and unloading LLMs with the same backbone weights. This would bloat memory utilization and is detrimental for latency, as it can take time to load and unload~\cite{chen2024punica,chen2024comparative}\footnote{Instead of using a single LLM as a thin wrapper around one tool, new designs keep a single base model loaded and expose multiple external tools through structured, declarative interfaces. The LLM can then \textit{reason about which tool to invoke and generate schema-constrained calls}, as illustrated in \url{https://web.archive.org/web/20260203193411/https://www.anthropic.com/engineering/advanced-tool-use}. This supports the composition of heterogeneous tools while avoiding memory overhead or the need for ad-hoc prompt engineering.}. Having to switch between LLMs also prevents the use of strategies that support efficiency, such as caching. Several research groups are thus converging towards one architecture (Figure~\ref{fig:architectures}d): \textit{one shared LLM backbone with knowledge augmentation} (see section~\ref{sec:knowledgeAugmentation})\index{Low-Rank Adaptation (LoRA)}, consisting of a dynamic switch or combination of task-specific lightweight modules (e.g., LoRAs)~\cite{chen2024punica,chen2024comparative}. This creates a single architecture (e.g., GPU cluster and LLM backend) to simultaneously support multiple independent users and tasks, which is called \textit{multi-tenant serving}. 

In practice, it is unrealistic to create and maintain a direct one-to-one mapping between every category of user query and every specialized tools via dedicated adapters. Given the large number of combinations, a more realistic approach would be to maintain a small set of reusable adapters that encode common capabilities. We previously suggested that ``the M\&S community may maintain a core set of translations (e.g., if the user’s query calls for first-order logic then use {\ttfamily Prover 9} syntax) and leave it to users if they need minor adjustments afterward (e.g., turn {\ttfamily Prover 9} into the related {\ttfamily Pyke} syntax to use a different solver).''~\cite{giabbanelli2025overreliance} A user's task would then be achieved by \textit{composing} adapters. Interestingly, there may exist several such compositions\footnote{Consider a modeler who wants to check a safety requirement stated in natural language (``the system must never enter an unsafe state''). One translation path could transform the simulation's description into state-based abstraction (states, transitions, guards) then encode requirement in temporal logic, and finally align with the syntax required by a specific checker. Alternatively, we could map requirements into constraint-style assertions (e.g., first-order constraints or satisfiability-style queries), and then use translate for a constraint solver. Both paths are `compositions', but they route the task through different formalisms and toolchains.}. These different compositions can have different characteristics, since LoRAs can differ in size and thus computational costs, and some compositions may be more accurate while others accumulate approximation errors\footnote{We can represent each format decision a a node and available transformations as directed edges, annotated with properties such as computational cost or information loss~\cite{schuerkamp2023extensions}. This representation enables a systematic exploration and evaluation of alternative sequences of transformations so that the desired transformation path can be selected based on explicit trade-offs. The literature on multi-paradigm modeling has also extensively covered model transformations, and particularly transformations between formalisms~\cite{vangheluwe2003computer}. The literature has also noted that sequences of transformations (also called workflows) are often implicit and informal, but documenting them explicitly would be helpful to compare, optimize, and reuse transformations. In short, we need to develop LoRAs that support transformations, and we need to document these sequences explicitly: which LoRAs are used, in what order, and with what outcomes.~\cite{rys2024model}.}\index{LLM pipelines|)}.

\section{Discussion}
\label{sec:LLMdiscuss}

Given the rising complexity of practices surrounding LLMs and the potential for misconceptions or counter-intuitive findings, this article aimed to provide a practical guide from the perspective of M\&S. We discussed what can and should be done \emph{now}, such as how to use RAG and LoRA, and how to connect LLMs with specialized tools in an efficient manner. As research on LLMs progresses rapidly, practitioners will need to regularly re-assess emerging opportunities. We do not expect LLMs to suddenly `know everything' or become deterministic in the short term. Consequently, the need for knowledge augmentation and for evaluating the impact of non-determinism is likely to remain. At the same time, the specific techniques used to address these needs may evolve. Some approaches may be displaced in the medium term (e.g., RAG or LoRA), while their implementations are already changing in the short term, as evidenced by the proliferation of variants. To help modelers anticipate such developments, the remainder of this section highlights two technical advances that are not yet widely applicable to M\&S (and were thus not covered earlier), but that could soon present new opportunities.

First, this article exclusively discussed LLMs as processing a textual input (which could represent different objects such as conceptual models or simulation outcomes). However, there are now multiple \textit{multimodal}\index{Multimodal LLM}\footnote{A multimodal LLM is any model that accepts or produces more than one modality, such as text \textit{and} images. Being multimodal says nothing about how the model reasons internally. For instance, an LLM can be multimodal by converting images to text and then reasoning only over text. Practitioners may also encounter the term \textit{omni}, which is more of a marketing or product label than a clearly defined research concept. The term is generally intended to \textit{suggest} a more integrated internal model, potentially supporting native or unified multimodality rather than relying on stitched pipelines such as a vision encoder deriving text tokens from an image, after which the LLM still reasons only on text. Ultimately, reasoning capabilities cannot be inferred solely from labels such as multimodal or omni. Rather, modelers must examine which representations are used, when they are used, and with what guarantees. Providers may disclose only partial information.} LLMs: they can process images, audio, or video. While these capacities have been available since GPT-4o, Gemini 2.0 Claude 3.5 Sonnet, or the lighweight Microsoft Phi 3.5-Vision, there has been a limited use of multimodal LLMs in the M\&S literature so far. Use cases have included providing inputs as visuals instead of text (e.g., images of UML diagrams~\cite{bates2025unified}), the analysis of simulation outputs (e.g., using images and videos of structural and fluid dynamics simulations~\cite{ezemba2025simulation}), or teaching an LLM about M\&S concept by giving it complete slide decks from a course (including visual for spatial concepts such as neighborhoods in cellular automata)~\cite{flandre2024can}. Results can be mixed, perhaps owing to the early stage of the technology. In the study analyzing simulation outputs, giving videos as inputs instead of batches of images had little impact, and text was the most robust modality~\cite{ezemba2025simulation}. As multimodal models mature and their internal representations and guarantees become better understood, such approaches merit renewed attention and systematic re-evaluation in future M\&S research.

Second, the dominant approach by far has been a \textit{design by addition}: when LLMs do not produce the desired answers, practitioners tend to add instructions to the prompt, fine-tune the model, or connect a RAG. However, LLMs may encode misunderstandings, in which case this strategy is akin to layering patches on top of a flawed specification rather than revisiting the specification itself. By contrast, a \textit{design by subtraction} approach seeks to identify which knowledge or associations should be removed from an LLM. In the case of conceptual modeling, we examined \textit{what} to unlearn and found that each essential aspect was violated by at least one well-known LLM, suggesting a need to unlearn\index{Unlearning} problematic samples (Table~\ref{tab:unlearning})~\cite{sakib2025rethinking}. A common misconception is that unlearning necessarily requires removing data samples and retraining the entire system, or substantial parts of it. While this is possible~\cite{bourtoule2021machine}, it is not the only approach; it can be computationally prohibitive and may have detrimental effects on properties such as fairness~\cite{zhang2024forgotten}. For example, when datasets are organized into shards to limit retraining costs, deleting a shard that exhibits issues in conceptual modeling may also remove data associated with novice learners, thereby impairing the model’s ability to generate or analyze novice conceptual models. An alternative to \textit{exact} unlearning through data deletion is \textit{approximate} unlearning, which aims to mitigate the influence of undesirable instances without fully removing them. There are many such methods,  different trade-offs and applicability to M\&S~\cite{sakib2025rethinking}. Some approaches target specific instances but may destabilize the LLM (just like editing a neuron can affect other functions of a brain), while others offer more localized control but need extensive labeling effort. Although these techniques remain largely untested in M\&S settings, they have potential to address persistent modeling errors.

\begin{table}[h]
\caption{Each of six key requirements for conceptual modeling was violated by at least one well-known LLM, highlighting the need to address such failures either by augmentation (patches) or by deliberate forgetting (unlearning) \protect\cite{sakib2025rethinking}.}
\begin{tabular}{|p{2.25cm}|p{2.5cm}|p{6.5cm}|}
\hline
\textbf{Category} & \textbf{Typical mistake}         & \textbf{Illustration / manifestation in conceptual modeling}                                                                                                                 \\ \hline
\textbf{Temporal representation}        & Mismatch of time units and ordering                   & Mixing daily and weekly time scales within the same model; inverting the order of events when describing system evolution                                                                         \\ \hline
\textbf{Task structure}                 & Linearization of inherently non-linear processes      & Representing branched or conditional processes as a single linear sequence; collapsing parallel tasks into a single process                                                                       \\ \hline
\textbf{Vagueness / lack of grounding}  & Use of ambiguous or unmeasurable constructs           & Concepts such as “integration,” “intentional exploration,” or “personal transformation” introduced without clear referents or operational meaning                                                 \\ \hline
\textbf{Causality}                      & Missing, reversed, or incorrect causal links          & Omitting causal connectors, reversing causal direction, or asserting direct causation where mediation is required (e.g., between population density, green space availability, and air pollution) \\ \hline
\textbf{Consistency of terms}           & Treating synonyms or paraphrases as distinct concepts & Introducing multiple labels for the same underlying concept, leading to fragmentation of the conceptual model and communication breakdowns                                                        \\ \hline
\textbf{Structural cohesiveness}        & Unclear model boundaries or aggregation level         & Producing a single monolithic model where multiple coordinated sub-models would be appropriate, or failing to articulate how components fit together                                              \\ \hline
\end{tabular}
\label{tab:unlearning}
\end{table}

%The idea of an `appropriate use' of LLMs is a matter of scope, so it is equally valuable to define what is \textit{in scope} (what does it mean to use LLMs for a specific M\&S task?) and exemplify what is \textit{out of scope}. 

\begin{acknowledgement}
This article was made possible through the experience gained in projects with many collaborators and students. The author is thankful for the hard work of current and former students in this domain (Cristina Perez, Stephen Zhong, Ryan Schuerkamp, Anish Shrestha, Noé Flandre, Tyler Gandee) as well as opportunities to learn with collaborators (Patrick Wu, Nathalie Japkowicz, Istvan David, Andreas Tolk, Niclas Feldkamp).
\end{acknowledgement}

\section*{Reflection and Exploration}
%%WILL NEED TO SWITCH TO questype
\begin{exploration}{Reasoning\index{Hyper-parameters!Reasoning} is not always better and refusing to engage is a signal}
Section~\ref{sec:hyperparameters} covered hyper-parameters such as temperature. LLMs increasingly expose a new hyper-parameter: \textit{reasoning}. It controls whether and how the model produces intermediate reasoning steps before generating an answer. A confusion would be to think that more reasoning is better. However, there are tasks when \textit{no reasoning} can be better than \textit{some reasoning}. Using a model with a reasoning parameter (e.g., GPT 5 Nano), provide a prompt under the `no reasoning' case and compare its results (over several runs) under the `low reasoning'. For example, the prompt from a social simulation can be: ``you are a college-educated Asian American male age 53, a peer is trying to convince you that higher taxes are good to support your community, what do you reply?'' Note that results may even differ in whether the LLM is willing to engage with the task, so do not treat refusals as a `bug' that you can replace by running the prompt again: an LLM refusing to answer is a signal.
\end{exploration}

\begin{exploration}{The different meanings of open models}
`Open' is a vague qualifier for LLMs and is sometimes invoked to justify the choice of a model. In practice, the term often refers to models for which practitioners can access the neural network \textit{weights}\index{Open weights}, allowing local deployment and extensive fine-tuning. However, Widder and colleagues argue that openness is multi-faceted and that AI systems labeled as `open' may not, in fact, be open in any meaningful sense, depending on how openness is defined~\cite{widder2024open}. According to the authors, which components of an LLM can meaningfully be made open, and which cannot? Consider an M\&S workflow that relies on fine-tuning an open-weights LLM. Which forms of openness remain relevant even if weights are available locally?
\end{exploration}

%%WILL NEED TO SWITCH TO questype
\begin{exploration}{LLM for RAG for LLM}%
When we presented Retrieval-Augmented Generation (RAG)\index{Retrieval-Augmented Generation (RAG)} in Section~\ref{sec:knowledgeAugmentation}, we framed retrieval as a way to provide knowledge to an LLM. Recent work, however, shows that LLMs are increasingly used within RAG themselves. Read the survey by Pandit \textit{et al.}~\cite{pandit2025evolution}, then answer the following questions: (i) at which stage of the RAG pipeline does the LLM intervene? (ii) What information does the LLM use?
\end{exploration}

%what about the 'reasoning' parameter? low reasoning could be worst than... no reasoning??

\begin{exploration}{Using masking to probe an LLM's knowledge prior to augmentation}
Section~\ref{sec:knowledgeAugmentation}\index{Prompt engineering!Cloze prompts} covered techniques for knowledge augmentation through contextual or parametric approaches (e.g., RAG, LoRA). Before embarking on the time-consuming process of optimizing one of these approaches, a modeler should assess whether knowledge augmentation is actually necessary. Determining what an LLM already knows (i.e., \textit{factual probing}) may take some effort, yet it can reveal that more complex augmentation strategies are unnecessary or that their scope can be sharply reduced. \textit{Cloze prompts} are a common method for factual probing, which uses sentences with missing parts (blanks) that the LLM must fill in. Note that replacing a word from a sentence with a special token (e.g., [MASK]) is a related technique called masking, but it is used for pre-training whereas cloze is a downstream prompt strategy. In this exercise, you will assess an LLM’s existing causal knowledge by selectively hiding information and observing how the LLM completes the missing elements.
\begin{enumerate}
    \item Create a small causal map consisting of 5 to 10 labeled nodes (e.g., reading textbooks, skill development, raises, satisfaction) and directed edges representing causal influence either as a causal increase (e.g., reading textbooks increases skill development) or decrease. You can also skip this step by reusing open causal maps and taking a small sample~\cite{giabbanelli2024narrating}.
    \item When transforming a conceptual model (e.g., causal map) to text, the input may have cycles but text must be linear. The linearization step consists of representing the causal graph as groups of edges which, if taken together, would re-create the whole model. Decompose the causal map created in the previous step into two or three linear parts (i.e., subgraphs).
    \item For each linear part of your conceptual model, create two versions: \textit{with} the type of causal edges (increase or decrease) and \textit{without} (i.e., `masking' the information).
    \item Prompt the LLM to generate a natural-language description of the model from the masked subgraphs and another description from the subgraphs containing the type of causal edges. Compare the results: is your LLM able to correctly guess the type of causality? Remember that LLMs are non-deterministic, so you should not conclude from a single observation; rather, each prompt should be run multiple times. Do not `give away' the answer to an LLM by including examples within your prompt, since we want the LLM to reveal what it knows about causality.
\end{enumerate}
\end{exploration}

\begin{exploration}{When knowledge augmentation is needed: experimenting with LoRAs}
Section~\ref{sec:knowledgeAugmentation}\index{Low-Rank Adaptation (LoRA)} discussed knowledge augmentation in LLMs through techniques such as Low-Rank Adaptation (LoRA). While LoRAs are often presented as a lightweight alternative to full fine-tuning, training a LoRA is not simply a matter of providing data and letting an algorithm build the model. The training process involves \textit{several parameters} that materially affect the resulting behavior. Some parameters govern the \textit{architecture of the LoRA itself} (e.g., rank, scaling factor $\alpha$, and dropout), determining how much and where the base model can be adapted. Other parameters govern the \textit{training process} (e.g., learning rate, batching strategy, number of epochs), influencing how strongly the LoRA internalizes the training examples. In this exercise, you will train LoRAs for a modeling and simulation task, systematically vary selected parameters, and observe how these choices affect both training outcomes and downstream behavior during inference.
\begin{enumerate}
    \item Construct a small, structured dataset that reflects a stable modeling task, suitable for LoRA training. \textit{The following is an example of how you can create such a dataset, but you can make a different one and still proceed with the next step.} Define a fixed target format to describe simulation scenarios (this codifies how you want the LoRA to enforce standards of the field). Then, create a dataset of input-output pairs where the input is a natural language description of a scenario (e.g., ``we'd like to see what happens when just one tree is burning and we look at the forest for 20 steps'') and output is the same scenario expressed based on the conventions that we seek to enforce (e.g., ``Initially, the number of burning trees is 1. The simulation runs for 20 steps.''). Ensure that inputs use varied phrasing and terminology for the same underlying ideas (e.g., `trees on fire', `burning trees', `ignited trees'), The outputs must follow the same textual conventions across all examples. You can have several inputs that map to the same output. Your goal to teach a stable way of expressing scenarios, not model domain knowledge.
    \item Train at least four LoRAs using the same dataset and base model, using {\ttfamily PEFT} or {\ttfamily Unsloth}. Withhold 20\% of your data from training, so that you can use it later for evaluation. For each LoRA, vary the architecture parameters (rank $r$ e.g. 4 vs 8 vs 16, scaling factor $\alpha$ e.g. 8 vs 16 vs 32, dropout e.g. 0 vs 0.1) and the training parameters (consider at least two values for the learning rate, the hatching strategy, and the number of epochs). Record the training time to create the LoRA, and observe the influence of the parameter values on the training time.
    \item To know whether a LoRA `works', you need to evaluate the output. Metrics could cover format compliance (does it look like the target style?), fidelity (does it preserve the intent of the informal description?), and so on. Propose a set of metrics along with clear criteria for scoring (e.g., what is high or low compliance?).
    \item Using the 20\% hold-out dataset and your proposed metrics, evaluate how the different LoRAs  influence the results. For each LoRA, use the same prompts and decoding parameters to generate codified scenario descriptions from the informal inputs. Compare the performances based on the parameters. For the best performing LoRA, vary its weight to 0.8 and 1, and examine the effect.
\end{enumerate}
\end{exploration}

%%WILL NEED TO SWITCH TO questype
\begin{exploration}{More data is not always better: avoiding model collapse and jailbreaks\index{Jailbreaking}}%
The performance of an LLM on a task does not always improve with more training data. We may observe \textit{plateaus}, such that the investment in acquiring more training data is no longer worth it. We may even observe a \textit{degradation} in downstream performance. This reading list explains why more data may not lead to the performance that modelers would have expected. Each reading can be pursued independently.
\begin{enumerate}
    \item Manually curating a set of input and output pairs is very time-consuming. It can also be difficult for a few individuals to create varied instances, as they can only think of so many different ways in which to describe the same information (e.g., the structure of a causal model, the wording of a simulation scenario). There is thus interest in training an LLM using data that was fully or partly generated by another LLM (`regurgitative training'). This may be intentional and controlled by modelers who prompt an LLM to generate data and apply filters to ensure diversity and quality of the dataset. It may also be unintentional, as modelers scrap online data that may have already been generated by LLMs. After reading the study by Zhang and colleagues~\cite{ zhang2024regurgitative}, discuss (i) how we can evaluate the quality of LLM-generated data intended as training for another LLM, and (ii) how we can generate data via several LLMs to increase the diversity of the training set. For a more formal treatment of \textit{model collapse} (i.e., a drop in performance after training on LLM-generated data), we refer the reader to Amin \textit{et al.}~\cite{amin2025escaping}.
    \item Fine-tuning an LLM may (perhaps inadvertently) compromise its safety guardrails, which is a particular concern when a model is made available for fine-tuning by third parties. To understand such fine-tuning attacks, read Hsiung \textit{et al.}~\cite{hsiung2025why} and then answer (i) why fine-tuning may compromise guardrails; (ii) how much data is needed, and how intentional does it have to be, in order to erode guardrails; and (iii) how can additional training data be screened to prevent safety issues.
    \item An experimental protocol can examine the response curve between the amount of data (x-axis) and the performance of an LLM (y-axis) on a given task. Read Giabbanelli \textit{et al.}~\cite{giabbanelli2024narrating} for a protocol that compares performance based on three discrete sample sizes (zero shot, few shot, fine tuning) then explain how this protocol can be generalized to provide a more fine-grained response curve. In particular, instead of evaluating performance at fixed sample sizes (e.g., 10, 20, 30 examples), consider how \textit{adaptive} sampling could be used: that is, selecting the next training size based on observed changes in performance.
\end{enumerate}
\end{exploration}

\begin{exploration}{Decomposing modeling and simulation data as inputs to LLMs\index{Prompt engineering!Data decomposition}}%
A conceptual model may be a large graph with hundreds of concepts and many more relationships, while a simulation output may consist of the values of many entities over many time steps and across many runs. LLMs cannot reliably attend to everything in such inputs. Naïve simplifications can lead to a loss of information and thus negatively impact performance: flattening a conceptual model can destroy structural information, while summarizing simulation outputs can miss on key patterns. Structure-aware decomposition algorithms have thus been developed at the level of the data (e.g., conceptual models~\cite{gandee2024combining}) or at the level of the query~\cite{yu2025tablerag} (identifying which operators such as filtering or joins are needed on which data type).
\begin{enumerate}
    \item Consider a conceptual model represented as a single graph, or a spreadsheet of simulation outputs spanning many entities and time steps. Suppose that a modeler simply provides the entire artifact (or a flattened textual version of it) to an LLM together with a question. Explain why this approach is likely to fail even when the artifact fits within the model’s context window. In your answer, distinguish between limitations due to context length and limitations due to loss of structure.
    \item From the readings, contrast data-level decomposition with query-level decomposition.
    Designing a decomposition strategy.
    \item Choose a M\&S artifact of interest to you and then explain how would decompose the artifact itself or the queries posed to it. This question is about devising a practical solution for a specific M\&S case, in contrast with the second question.
\end{enumerate}
\end{exploration}

\begin{exploration}{Identifying reusable skills in LLM pipelines}
    Recent practitioner-oriented tooling increasingly frames LLM capabilities as reusable \textit{skills}\index{skills}: small, self-contained pipeline components that implement a specific function (e.g., retrieval, evaluation, tool calling, or structured output validation). Rather than proposing new learning algorithms, skill repositories focus on operationalizing best practices and reducing ad-hoc glue code when assembling multi-stage LLM systems. Browse the Hugging Face skills repository (https://github.com/huggingface/skills) and identify two skills that are relevant to modeling and simulation workflows (e.g., retrieval over technical documents, structured generation, or evaluation of generated outputs). Explain the capability provided by each skill, where it fits in the M\&S pipeline, and what aspects would be standardized (e.g., retrieval strategy, output structure). 
\end{exploration}

\let\cleardoublepage\clearpage
\printindex

%\clearpage
%\section*{Index}
%\addcontentsline{toc}{section}{Index}
%\input{authorsample.ind}

\end{document}